\documentclass[conference]{IEEEtran}
\usepackage{times}

\usepackage[numbers]{natbib}
\usepackage{multicol}
\usepackage[bookmarks=true]{hyperref}
\usepackage[algoruled,boxed,lined]{algorithm2e}
\usepackage{amsmath, amssymb}
\usepackage{graphicx}
\usepackage{subfigure}

\IEEEoverridecommandlockouts 
\begin{document}

\title{Emergent Real-World Robotic Skills via Unsupervised Off-Policy Reinforcement Learning}

% You will get a Paper-ID when submitting a pdf file to the conference system
\author{\thanks{\hrulefill}Archit Sharma$^\alpha$ \thanks{$\alpha$ Work done in Google AI residency}, Michael Ahn, Sergey Levine, Vikash Kumar, Karol Hausman$^\beta$, Shixiang Gu$^\beta$  \thanks{$\beta$ equal contribution}\\\thanks{Contact: architsh@google.com}
Google Research}

% \author{\authorblockN{Michael Shell}
% \authorblockA{School of Electrical and\\Computer Engineering\\
% Georgia Institute of Technology\\
% Atlanta, Georgia 30332--0250\\
% Email: mshell@ece.gatech.edu}
% \and
% \authorblockN{Homer Simpson}
% \authorblockA{Twentieth Century Fox\\
% Email: homer@thesimpsons.com}
% \and
% \authorblockN{James Kirk\\ and Montgomery Scott}
% \authorblockA{Starfleet Academy\\
% San Francisco, California 96678-2391\\
% Telephone: (800) 555--1212\\

% avoiding spaces at the end of the author lines is not a problem with
% conference papers because we don't use \thanks or \IEEEmembership

%Homer Simpson\authorrefmark{2},
%Montgomery Scott\authorrefmark{3} and
%Eldon Tyrell\authorrefmark{4}}
%Telephone: (800) 555--1212, Fax: (888) 555--1212}
%\authorblockA{\authorrefmark{4}Tyrell Inc., 123 Replicant Street, Los Angeles, California 90210--4321}}

\maketitle

\begin{abstract}
Reinforcement learning provides a general framework for learning robotic skills while minimizing engineering effort. However, most reinforcement learning algorithms assume that a well-designed reward function is provided, and learn a single behavior for that single reward function. Such reward functions can be difficult to design in practice. Can we instead develop efficient reinforcement learning methods that acquire diverse skills without any reward function, and then re-purpose these skills for downstream tasks? In this paper, we demonstrate that a recently proposed unsupervised skill discovery algorithm can be extended into an efficient off-policy method, making it suitable for performing unsupervised reinforcement learning in the real world. Firstly, we show that our proposed algorithm provides substantial improvement in learning efficiency, making reward-free real-world training feasible. Secondly, we move beyond the simulation environments and evaluate the algorithm on real physical hardware. On quadrupeds, we observe that locomotion skills with diverse gaits and different orientations emerge without any rewards or demonstrations. We also demonstrate that the learned skills can be composed using model predictive control for goal-oriented navigation, without any additional training.
\end{abstract}

\IEEEpeerreviewmaketitle

% AS add more citations for relevant papers
\section{Introduction}
\label{sec:introduction}
% Intro version 2 by Karol
Reinforcement learning (RL) has the potential of enabling autonomous agents to exhibit intricate behaviors and solve complex tasks from high-dimensional sensory input without hand-engineered policies or features~\citep{sutton1998introduction,riedmiller2005neural,mnih2013playing,lillicrap2015continuous,gu2016continuous}.
These properties make this family of algorithms particularly applicable to the field of robotics where hand-engineering features and control policies have proven to be challenging and difficult to scale~\citep{kohl2004policy,kober2009policy,riedmiller2009reinforcement,kober2013reinforcement,gu2017deep,kalashnikov2018qt}.
However, applying RL to real-world robotic problems has not fully delivered on its promise.
One of the reasons for this is that the assumptions that are required in a standard RL formulation are not fully compatible with the requirements of real-world robotics systems. 
One of these assumptions is the existence of a ground truth reward signal, provided as part of the task. 
While this is easy in simulation, in the real world this often requires special instrumentation of the setup, as well as the ability to reset the environment after every learning episode, which often requires tailored reset mechanisms or manual labor. 
If we could relax some of these assumptions, we may be able to fully utilize the potential of RL algorithms in real-world robotic problems.%Can we relax some of these assumptions to fully utilize the potential of RL algorithms in real-world robotics?

\begin{figure}[!htb]
    \centering
    \includegraphics[width=0.48\textwidth]{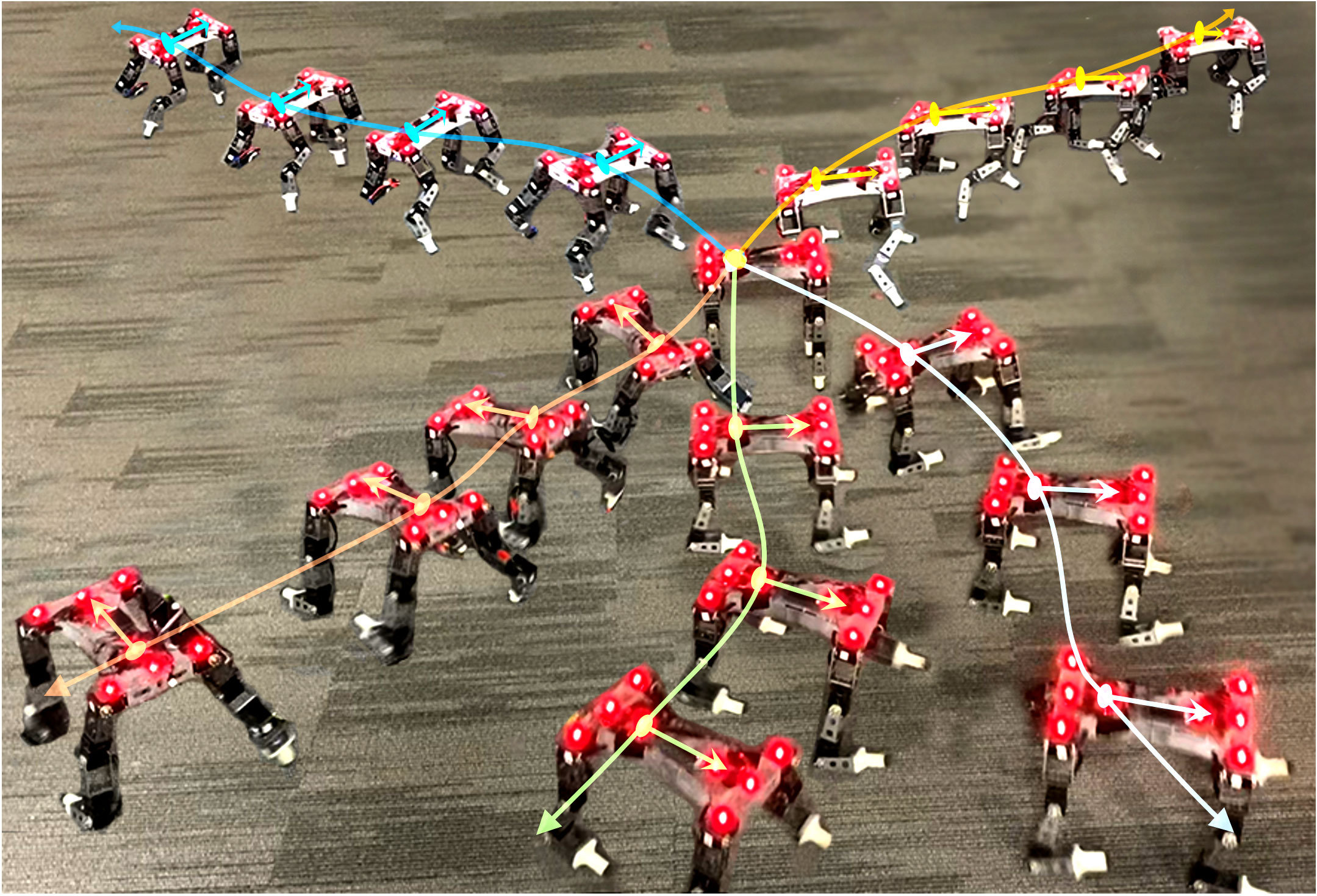}
    \caption{\small A 12 degree of freedom quadruped (D'Kitty) discovers diverse locomotion skills without any rewards or demonstrations. We develop off-DADS, an asynchronous and off-policy version of Dynamics-Aware Discovery of Skills (DADS) \citep{sharma2019dynamics}, that enables sample-efficient skill discovery based on mutual-information based optimization.}
    \label{fig:superposed_skills}
\end{figure}

% In this work, we attempt to take a step towards lifting one of these limiting assumptions - the presence of task-specific reward functions. 
% A method that can learn skills without any external reward supervision can then re-purpose those skills to solve downstream tasks using only a limited amount of interaction.
In this context for robotics, the recent work in unsupervised learning becomes relevant --- we can learn skills without any external reward supervision and then re-purpose those skills to solve downstream tasks using only a limited amount of interaction. 
Of course, when learning skills without any reward supervision, we have limited control over the kinds of skills that emerge. 
Therefore, it is critical for unsupervised skill learning frameworks to optimize for diversity, so as to produce a large enough repertoire of skills such that potentially useful skills are likely to be part of this repertoire.
In addition, a framework like this needs to offer the user some degree of control over the dimensions along which the algorithm explores.
Prior works in unsupervised reinforcement learning~\citep{oudeyer2009intrinsic,mohamed2015variational,pathakICMl17curiosity,achiam2018variational, eysenbach2018diversity, florensa2017stochastic, sharma2019dynamics} have demonstrated that interesting behaviors can emerge from reward-free interaction between the agent and environment. 
In particular, \citep{eysenbach2018diversity, florensa2017stochastic, sharma2019dynamics} demonstrate that the skills learned from such unsupervised interaction can be harnessed to solve downstream tasks. 
However, due to their sample-inefficiency, these prior works in unsupervised skill learning have been restricted to simulation environments (with a few exceptions such as~\citet{baranes2013active,pong2019skew,lee2019efficient}) and their feasibility of executing on real robots remains unexplored.
%%SL.1.31: the goal conditioned work has demonstrated more efficient learning (e.g., Skew-Fit), with results on real-world tasks. Also make sure we didn't miss any work from Pierre-Yves Oudeyer that might do something like this in the real world too. Also Lisa's SMM paper.

In this paper, we address the limiting sample-inefficiency challenges of previous reward-free, mutual-information-based learning methods and demonstrate that it is indeed feasible to carry out unsupervised reinforcement learning for acquisition of robotic skills.
%%SL.1.31: see my comment before -- there is prior work that does unsupervised RL with real robots
To this end, we build on the work of~\citet{sharma2019dynamics} and derive a sample-efficient, off-policy version of a mutual-information-based, reward-free RL algorithm, Dynamics-Aware Discovery of Skills (DADS), which we refer to as off-DADS. 
Our method uses a mutual-information-based objective for diversity of skills and specification of task-relevant dimensions (such as x-y location of the robot) to specify where to explore. 
Moreover, we extend off-DADS to be able to efficiently collect data on multiple robots, which together with the efficient, off-policy nature of the algorithm, makes reward-free real-world robotics training feasible.
We evaluate the asynchronous off-DADS method on D'Kitty, a compact cost-effective quadruped, from the ROBEL robotic benchmark suite~\citep{ahn2019robel}. 
We demonstrate that diverse skills with different gaits and navigational properties can emerge, without any reward or demonstration. 
We present simulation experiments that indicate that our off-policy algorithm is up to 4x more efficient than its predecessor.
In addition, we conduct real-world experiments showing that the learned skills can be harnessed to solve downstream tasks using model-based control, as presented in~\citep{sharma2019dynamics}.

\section{Related Work}
\label{sec:related_work}

%SG.01.27: TODO: add more works from before 2016

Our work builds on a number of recent works~\citep{gu2017deep,kumar2016optimal,kalashnikov2018qt,mahmood2018benchmarking,haarnoja2018learning,nagabandi2019deep} that study end-to-end reinforcement learning of neural policies on real-world robot hardware, which poses significant challenges such as sample-efficiency, reward engineering and measurements, resets, and safety~\citep{gu2017deep,dulac2019challenges,zhu2020ingradients}. \citet{gu2017deep,kalashnikov2018qt,haarnoja2018learning,nagabandi2019deep} demonstrate that existing off-policy and model-based algorithms are sample efficient enough for real world training
%%SL.1.31: "Nearing" seems like the wrong word -- if they are successful, then they're not "nearing," they are already there :)
%SG.01.31 agree
of simple manipulation and locomotion skills given reasonable task rewards. \citet{eysenbach2017leave,zhu2020ingradients} propose reset-free continual learning algorithms and demonstrate initial successes in simulated and real environments. To enable efficient reward-free discovery of skills, our work aims to address the sample-efficiency and reward-free learning jointly through a novel off-policy learning framework. %VK: There are frequent jumps between sample-efficiency, reset-free, reward-free points in this paragraph. It will be good to harmonize them a little.

%%SL.1.31: Overall, I think we are likely to get into trouble with this pitch. Our message here is that basically "reward design is hard, so let's learn without rewards." But this is likely to come off as very strange to reviewers, especially in robotics, who will say "well, yeah, but if you learn without rewards, how can you learn the right thing?" I don't think we can just completely ignore this issue, it will cause people to think we are overselling. We will need to acknowledge it. So we could consider (here or more likely in intro) saying something like this: Of course, when learning skills without any reward supervision, we have limited control over the kinds of skills that emerge. Therefore, it is critical for unsupervised skill learning frameworks to optimize for diversity, so as to produce a large enough repertoire of skills such that potentially useful skills are likely to be part of this repertoire, and to offer the user some degree of control over the dimensions along which the algorithm explores." [or something like that], and then discuss our method does this, using MI for diversity and using specification of task-relevant dimensions (i.e., xy) to specify where to explore.

Reward engineering has been a major bottleneck not only in robotics, but also in general RL domains. There are two kinds of approaches to alleviate this problem. The first kind involves recovering a task-specific reward function with alternative forms of specifications, such as inverse RL~\citep{ng2000algorithms,abbeel2004apprenticeship,ziebart2008maximum,hadfield2017inverse} or preference feedback~\citep{christiano2017deep}; however, these approaches still require non-trivial human effort. The second kind proposes an intrinsic motivation reward that can be applied to different MDPs to discover useful policies, such as curiosity for novelty~\citep{schmidhuber1991curious,oudeyer2009intrinsic,schmidhuber2010formal,bellemare2016unifying,pathak2017curiosity,colas2018curious}, entropy maximization~\citep{hazan2018provably,pong2019skew,lee2019efficient,ghasemipour2019divergence}, and mutual information~\citep{klyubin2005all,jung2011empowerment,daniel2012hierarchical,florensa2017stochastic,eysenbach2018diversity,mohamed2015variational,sharma2019dynamics}. Ours extends the dynamics-based mutual-information objective from~\citet{sharma2019dynamics} to sample-efficient off-policy learning.
%%SL.1.31: I still don't think we should call it empowerment (because people who work on empowerment will say it's not the same).
%SG.01.31: replaced with mutual information 

Off-policy extension to DADS~\citep{sharma2019dynamics} poses challenges beyond those in standard RL~\citep{precup2000eligibility,jiang2015doubly,thomas2016data,munos2016safe}. Since we learn an action abstraction that can be related to a low-level policy in hierarchical RL (HRL)~\citep{sutton1999between,stolle2002learning,konidaris2011autonomous,bacon2017option,nachum2018data}, we encounter similar difficulties as in off-policy HRL~\citep{nachum2018data,levy2017learning}. We took inspirations from the techniques introduced in~\citep{nachum2018data} for stable off-policy learning of a high-level policy; however, on top of the non-stationarity in policy, we also need to deal with the non-stationarity in reward function as our DADS rewards are continually updated during policy learning. We successfully derive a novel off-policy variant of DADS that exhibits stable and sample-efficient learning.

\section{Background}
\label{sec:background}
In this section, we setup the notation to formally introduce the reinforcement learning problem and the algorithmic foundations of our proposed approach. We work in a Markov decision process (MDP) $\mathcal{M} = (\mathcal{S}, \mathcal{A}, p, r)$, where $\mathcal{S}$ denotes the state space of the agent, $\mathcal{A}$ denotes the action space of the agent, $p : \mathcal{S} \times \mathcal{S} \times \mathcal{A} \rightarrow [0, \infty)$ denotes the underlying (stochastic) dynamics of the agent-environment which can be sampled starting from the initial state distribution $p_0: \mathcal{S} \rightarrow [0, \infty)$, and a reward function $r: \mathcal{S} \times \mathcal{A} \rightarrow [0, \infty)$. The goal of the optimization problem is to learn a controller $\pi(a_t \mid s_t)$  which maximizes $\mathbb{E}[\sum_t \gamma^tr(s_t, a_t)]$ for a discount factor $\gamma \in [0, 1)$.

% add citations for relevant methods
Within deep reinforcement learning, there are several methods to optimize this objective. In particular, off-policy methods centered around learning a Q-function \citep{lillicrap2015continuous, haarnoja2018soft, gu2016continuous,kalashnikov2018qt} are known to be suitable for reinforcement learning on robots. At a high level, algorithms estimate ${Q^\pi(s_t, a_t) = \mathbb{E}[\sum_{i\geq t} \gamma^{i-t} r(s_i, a_i)]}$, where the expectation is taken over state-action trajectories generated by the executing policy $\pi$ in the MDP $\mathcal{M}$ after taking action $a_t$ in the state $s_t$. Crucially, $Q^\pi$ can be estimated using data collected from arbitrary policies using the temporal-difference learning (hence off-policy learning). For continuous or large discrete action spaces, a parametric policy can be updated to ${\pi'(a \mid s) \leftarrow \textrm{argmax}_a Q^\pi(s, a)}$, which can be done approximately using stochastic gradient descent when $Q$ is differentiable with respect to $a$ \citep{lillicrap2015continuous, haarnoja2018softapp, gu2017interpolated}. While the off-policy methods differ in specifics of each step, they alternate between estimating $Q^\pi$ and updating $\pi$ using the $Q^\pi$ till convergence. The ability to use trajectories sampled from arbitrary policies enables these algorithms to be sample efficient.

\subsection{Unsupervised Reinforcement Learning}
In an unsupervised learning setup, we assume a MDP ${\mathcal{M} = (\mathcal{S}, \mathcal{A}, p)}$ without any reward function $r$, retaining the previous definitions and notations. The objective is to systematically acquire diverse set of behaviors using autonomous exploration, which can subsequently be used to solve downstream tasks efficiently. To this end, a skill space $\mathcal{Z}$ is defined such that a behavior $z \in \mathcal{Z}$ is defined by the policy $\pi(a|s, z)$. To learn these behaviors in a reward-free setting, the information theoretic concept of mutual information is generally employed. Intuitively, mutual information ${\mathcal{I}(x, y)}$ between two random variables $x, y$ is high when given $x$, the uncertainty in value of $y$ is low and vice-versa. Formally, $$\mathcal{I}(x, y) = \int p(x, y) \log \frac{p(x, y)}{p(x)p(y)} dx dy$$

Dynamics-aware Discovery of Skills (DADS) \cite{sharma2019dynamics} uses the concept of mutual information to encourage skill discovery with predictable consequences. It uses the following conditional mutual information formulation to motivate the algorithm: $\mathcal{I}(s', z \mid s)$ where $s'$ denotes the next state observed after executing the behavior $z$ from the state $s$. The joint distribution can be factorized as follows: ${p(z, s, s') = p(z)p(s \mid z) p(s' \mid s, z)}$, where $p(z)$ denotes the prior distribution over $\mathcal{Z}$, $p(s \mid z)$ denotes the stationary distribution induced by $\pi(a \mid s, z)$ under the MDP $\mathcal{M}$ and $p(s' \mid s, z) = \int p(s' \mid s, a) \pi(a \mid s,z) da$ denotes the transition dynamics. The conditional mutual information can be written as $$ \mathcal{I}(s', z \mid s) = \int p(z, s, s') \log \frac{p(s' \mid s, z)}{p(s' \mid s)} dz ds ds'$$
At a high level, optimizing for $\mathcal{I}(s', z \mid s)$ encourages $\pi$ to generate trajectories such that $s'$ can be determined from $s, z$ \textit{(predictability)} and simultaneously encourages $\pi$ to generate trajectories where $s'$ cannot be determined well from $s$ without $z$ \textit{(diversity)}. Note, computing $\mathcal{I}$ is intractable due to intractability of $p(s' \mid s, z)$ and $p(s' \mid s)$. However, one can motivate the following reinforcement learning maximization for $\pi$ using variational inequalities and approximations as discussed in \cite{sharma2019dynamics}:
$$J(\pi) = \mathbb{E}_{z, s, s' \sim p(z, s, s')}[r(s, z, s')]$$
$$r(s, z, s') = \frac{q_\phi(s' \mid s, z)}{\sum_{i=1}^L q_\phi(s' \mid s, z_i)} + \log L$$
for $\{z_i\}_{i=1}^L\sim p(z)$ where $q_\phi$ maximizes $$J(q_\phi) = \mathbb{E}_{ z,s,s' \sim p(z, s, s')}[\log{q_\phi(s' \mid s, z)}]$$
\citet{sharma2019dynamics} propose an on-policy alternating optimization: At iteration $t$, collect a batch $\mathcal{B}^{(t)}$ of trajectories from the current policy $\pi^{(t)}$ to simulate samples from $p(z, s, s')$, update $q_\phi^{(t)}\rightarrow q_\phi^{(t+1)}$ on $\mathcal{B}^{(t)}$ using stochastic gradient descent to approximately maximize $J(q_\phi)$, label the transitions with reward $r^{(t+1)}(s, z, s')$ and  update $\pi^{(t)} \rightarrow \pi^{(t+1)}$ on $\mathcal{B}^{(t)}$ using any reinforcement learning algorithm to approximately maximize $J(\pi)$. Note, the optimization encourages the policy $\pi$ to produce behaviors predictable under $q_\phi(s' | s, z)$, while rewarding the policy for producing diverse behaviors for different $z \in \mathcal{Z}$. This can be seen from the definition of $r(s, z, s')$: The numerator will be high when the transition $s \rightarrow s'$ has a high log probability under the current skill $z$ (high $q_\phi(s' \mid s, z)$ implies high predictability), while the denominator will be lower if the transition has low probability under $z_i$ (low $q_\phi(s' \mid s, z_i)$ implies $q_\phi$ is expecting a different transition under the skill $z_i$).

Interestingly, the variational approximation $q_\phi(s' \mid s, z)$, called skill dynamics, can be used for model-predictive control. Given a reward function at test-time, the sequence of skill $z \in Z$ can be determined online using model-predictive control by simulating trajectories using skill dynamics $q_\phi$.

\section{Towards Real-world Unsupervised Learning}
The broad goal of this section is to motivate and present the algorithmic choices required for accomplishing reward-free reinforcement learning in the real-world. We address the issue of sample-efficiency of learning algorithms, which is the main bottleneck to running the current unsupervised learning algorithms in the real-world. In the same vein, an asynchronous data-collection setup with multiple actors can substantially accelerate the real-world execution. We exploit the off-policy learning enterprise to demonstrate unsupervised learning in the real world, which allows for both sample-efficient and asynchronous data collection through multiple actors ~\cite{kalashnikov2018qt}.
\subsection{Off-Policy Training of DADS}
\label{sec:method}

We develop the off-policy variant of DADS, which we call off-DADS. For clarity, we can restate $J(\pi)$ in the more conventional form of expected discounted sum of rewards. Using the definition of the stationary distribution $p(s \mid z) = \sum_{t=0}^T\gamma^t p(s_t = s \mid z)$ for a $\gamma$-discounted episodic setting of horizon $T$, we can write:
$$J(\pi) = \mathbb{E}[\sum_{t=0}^{T-1} \gamma^{t}r(s_t, z, s_{t+1})]$$
where the expectation has been taken with respect to trajectories generated by $\pi(a \mid s, z)$ for $z \sim p(z)$. This has been explicitly shown in Appendix~\ref{app:stationary}. Now, we can write the $Q$-value function as 
$$Q^\pi(s_t, a_t) = \mathbb{E}[\sum_{i \geq t} \gamma^{i-t}r(s_t, z, s_{t+1})]$$
For problems with a fixed reward function, we can use off-the-shelf off-policy reinforcement learning algorithms like soft actor-critic \citep{haarnoja2018soft, haarnoja2018softapp} or deep deterministic policy gradient \citep{lillicrap2015continuous}. At a high level, we use the current policy $\pi^{(t)}$ to sample a sequence of transitions from the environment and add it to the replay buffer $\mathcal{R}$. We uniformly sample a batch of transitions $\mathcal{B}^{(t)} = \{(s_i, z_i, a_i, s'_i)\}_{i=0}^B$ from $\mathcal{R}$ and use it to update $\pi^{(t)}$ and $Q^{\pi^{(t)}}$.
%%SL.1.31: For the above paragraph, I feel like it could be written with a bit more motivation for the method (i.e., explain why these things are done in this way, rather than just stating that they are)

However, in this setup: (a) the reward is non-stationary as $r(s, z, s')$ depends upon $q_\phi$, which is learned simultaneously to $\pi, Q^\pi$ and (b)
learning $q_\phi$ involves maximizing $J(q_\phi)$ which implicitly relies on the current policy $\pi$ and the induced stationary distribution $p(s \mid z)$. For (a), we recompute the reward $r(s, z, s')$ for the batch $\mathcal{B}^{(t)}$ using the current iterate $q_\phi^{(t)}$. For (b), we propose two alternative methods:
\begin{itemize}
    \item We use samples from current policy $\pi^{(t)}$ to maximize $J(q_\phi)$. While this does not introduce any additional bias, it does not take advantage of the off-policy data available in the replay buffer.
    \item Reuse off-policy data while maximizing $J(q_\phi)$.
\end{itemize}
\begin{algorithm}
    \label{algorithm:off_dads}
    \SetAlgoLined
    Initialize parameters $\pi(a \mid s, z), q_\phi(s' \mid s, z)$\;
    Initialize replay buffer $\mathcal{R}$\;
    // collector threads \\
    \While{not done}{
        Sync $\pi_c \gets \pi$ \;
        Sample $z \sim p(z)$\;
        Collect episode using $\pi_c$; // store $\pi_c(a \mid s, z)$ \\
    }
    // training thread \\
    $s \gets 0, n \gets 0$\;
    \While{not done}{
        \While{$n < s + newsteps$}{
            Sync $\mathcal{B}$; $n += size(\mathcal{B})$; // get data from actors\\
            $\mathcal{R} \gets \mathcal{R} \cup \mathcal{B}$\;
        }
        \For{$i\gets1$ \KwTo $T_q$}{
            Sample $\{s_j, z_j, a_j, s'_j, \pi_c(a_j \mid s_j, z_j)\}_{j=1}^{B_q} \sim \mathcal{R}$\;
            $w_j \gets clip(\frac{\pi(a_j \mid s_j,z_j)}{\pi_c(a_j \mid s_j, z_j)}, \frac{1}{\alpha}, \alpha)$\;
            Update $q_\phi$ using $\{s_j, z_j, s'_j, w_j)\}_{j=1}^{B_q}$\;
        }

        \For{$i\gets1$ \KwTo $T_\pi$}{
            Sample $\{s_j, z_j, a_j, s'_j, \pi_c(a_j \mid s_j, z_j)\}_{j=1}^{B_\pi} \sim \mathcal{R}$\;
            $r_j \gets r(s_j, z_j, s'_j)$; // DADS reward \\
            Update $\pi$ using $\{s_j, z_j, a_j, s'_j, r_j)\}_{j=1}^{B_\pi}$\;
        }
        $s\gets n$\;
    }
    \caption{Asynchronous off-DADS}
\end{algorithm}
To re-use off policy data for learning $q_\phi$, we have to consider importance sampling corrections, as the data has been sampled from a different distribution. While we can derive an unbiased gradient estimator, as discussed in Appendix~\ref{app:importance_sampling}, we motivate an alternate estimator which is simpler and more stable numerically, albeit biased. Consider the definition of $J(q_\phi)$:
\begin{align*}
    J(q_\phi) &= \mathbb{E}_{z, s, s' \sim p}\big[\log q_\phi(s' \mid s,z) \big]\\
    &= \int p(z)p(s \mid z) p(s' \mid s,z) \log q_\phi(s' \mid s,z) dz ds ds'\\
    &= \int p(z) p(s \mid z) \pi(a \mid s, z) p(s'\mid s, a) \\
    & \qquad\qquad\qquad\qquad \log q_\phi(s' \mid s, z) dz ds da ds'
\end{align*}
where we have used the fact that ${p(s' \mid s, z) = \int p(s' \mid s, a) \pi(a \mid s, z) da}$. Now, consider that the samples have been generated by a behavior policy $\pi_c(a \mid s, z)$. The corresponding generating distribution can be written as: ${p^{\pi_c}(z, s, s') = \int p(z) p_c(s \mid z) \pi(a \mid s, z) p(s' \mid s, a) da}$, where the prior $p(z)$ over $\mathcal{Z}$ and the dynamics $p(s'\mid s, a)$ are shared across all policies, and $p_c(s \mid z)$ denotes the stationary state distribution induced by $\pi_c$. We can rewrite $J(q_\phi)$ as
\begin{multline*}
    J(q_\phi) = \int p(z) p_c(s \mid z) \pi_c(a \mid s, z) p(s' \mid s, a) \\
                \frac{p(s \mid z) \pi(a \mid s, z)}{p_c(s \mid z) \pi_c(a \mid s, z)} \log q_\phi(s' \mid s, z) dz ds da ds'
\end{multline*}
which is equivalent to 
$$J(q_\phi) = \mathbb{E}_{z, s, a, s' \sim p^{\pi_c}}\Big[\frac{p(s \mid z) \pi(a \mid s, z)}{p_c(s \mid z) \pi_c(a \mid s, z)} \log q_\phi(s' \mid s, z)\Big]$$
Thus, the gradient for $J(q_\phi)$ with respect to $\phi$ can be written as:
\begin{align*}
    \nabla_\phi J(q_\phi) &= \mathbb{E}\Big[\frac{p(s \mid z) \pi(a \mid s, z)}{p_c(s \mid z) \pi_c(a \mid s, z)} \nabla_\phi\log q_\phi(s' \mid s, z)\Big]\\
    &\approx \frac{1}{B_q}\sum_{i=1}^{B_q}\Big[\frac{\pi(a_i \mid s_i, z_i)}{\pi_c(a_i \mid s_i, z_i)} \nabla_\phi\log q_\phi(s'_i \mid s_i, z_i)\Big]
\end{align*}

The estimator is biased because we compute the importance sampling correction as $w_i =clip(\frac{\pi(a_i \mid s_i, z_i)}{\pi_c(a_i \mid s_i, z_i)}, \frac{1}{\alpha}, \alpha)$ which ignores the intractable state-distribution correction $\frac{p(s \mid z)}{p_c (s \mid z)}$~\citep{gu2017interpolated}. This considerably simplifies the estimator while keeping the estimator numerically stable (enhanced by clipping) as compared to the unbiased estimator derived in Appendix~\ref{app:importance_sampling}. In context of off-policy learning, the bias due to state-distribution shift can be reduced using a shorter replay buffer.

Our final proposed algorithm is summarized in the Algorithm~\ref{algorithm:off_dads}. At a high level, we use $n$ actors in the environment which use the latest copy of the policy to collect episodic data. The centralized training script keeps adding new episodes to the shared replay buffer $\mathcal{R}$. When a certain threshold of new experience has been added to $\mathcal{R}$, the buffer is uniformly sampled to train $q_\phi$ to maximize $J(q_\phi)$. To update $\pi$, we sample the buffer uniformly again and compute $r(s, z, s')$ for all the transitions using the latest $q_\phi$. The labelled transitions can then be passed to any off-the-shelf off-policy reinforcement learning algorithm to update $\pi$ and $Q^\pi$.
\section{Experiments}
In this section, we experimentally evaluate our robotic learning method, off-DADS, for unsupervised skill discovery. First, we evaluate the off-DADS algorithm itself in isolation, on a set of standard benchmark tasks, to understand the gains in sample efficiency when compared to DADS proposed in \cite{sharma2019dynamics}, while ablating the role of hyperparameters and variants of off-DADS. Then, we evaluate our robotic learning method on D'Kitty from ROBEL \cite{ahn2019robel}, a real-world robotic benchmark suite. We also provide preliminary results on D'Claw from ROBEL, a manipulation oriented robotic setup in Appendix~\ref{appendix: D'Claw}.

\subsection{Benchmarking off-DADS}
\begin{figure}[!htb]
    \centering
    \begin{minipage}{0.24\textwidth}
    \centering
    \small (a) Half-Cheetah\par
    \includegraphics[width=\textwidth]{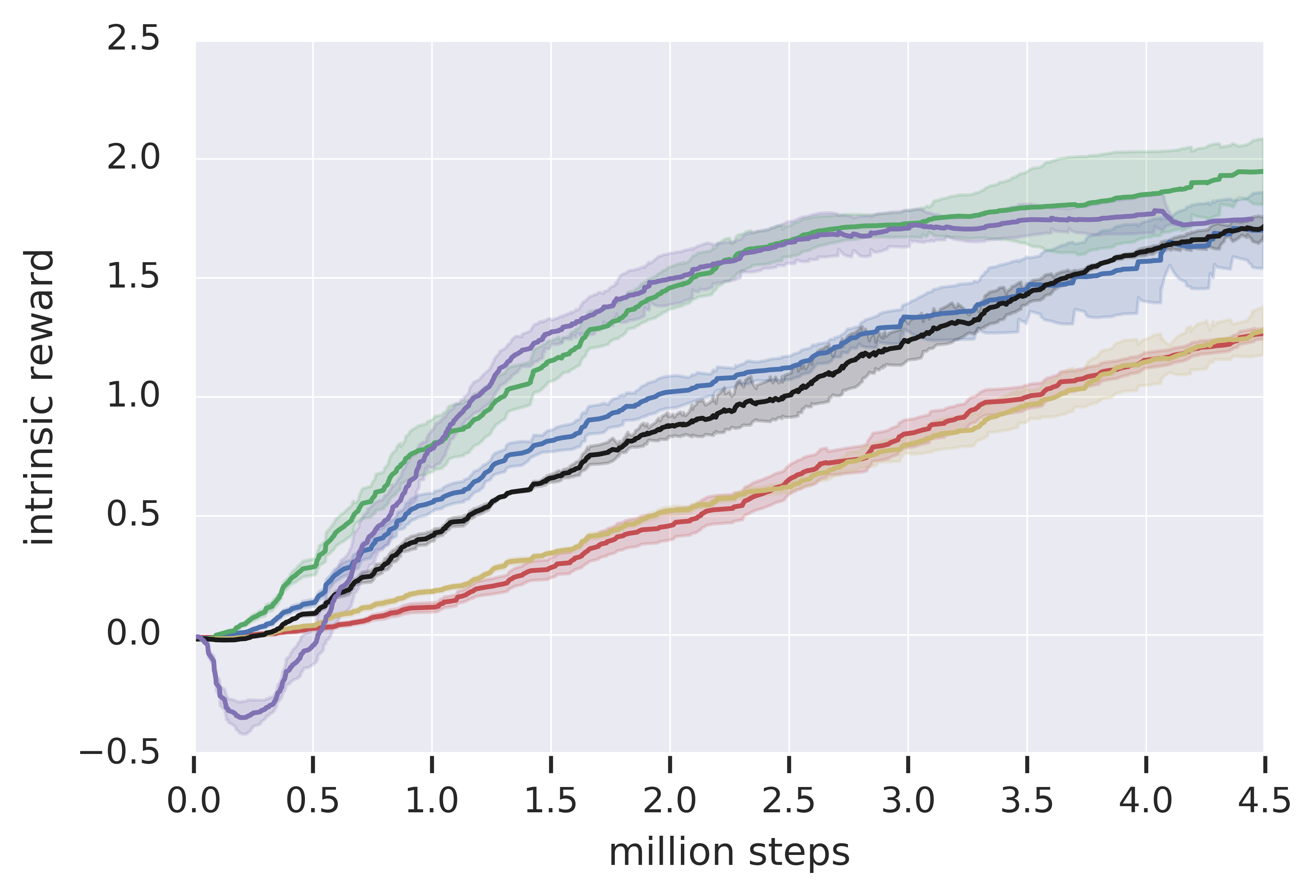}
    \small (b) Ant  \par
    \includegraphics[width=\textwidth]{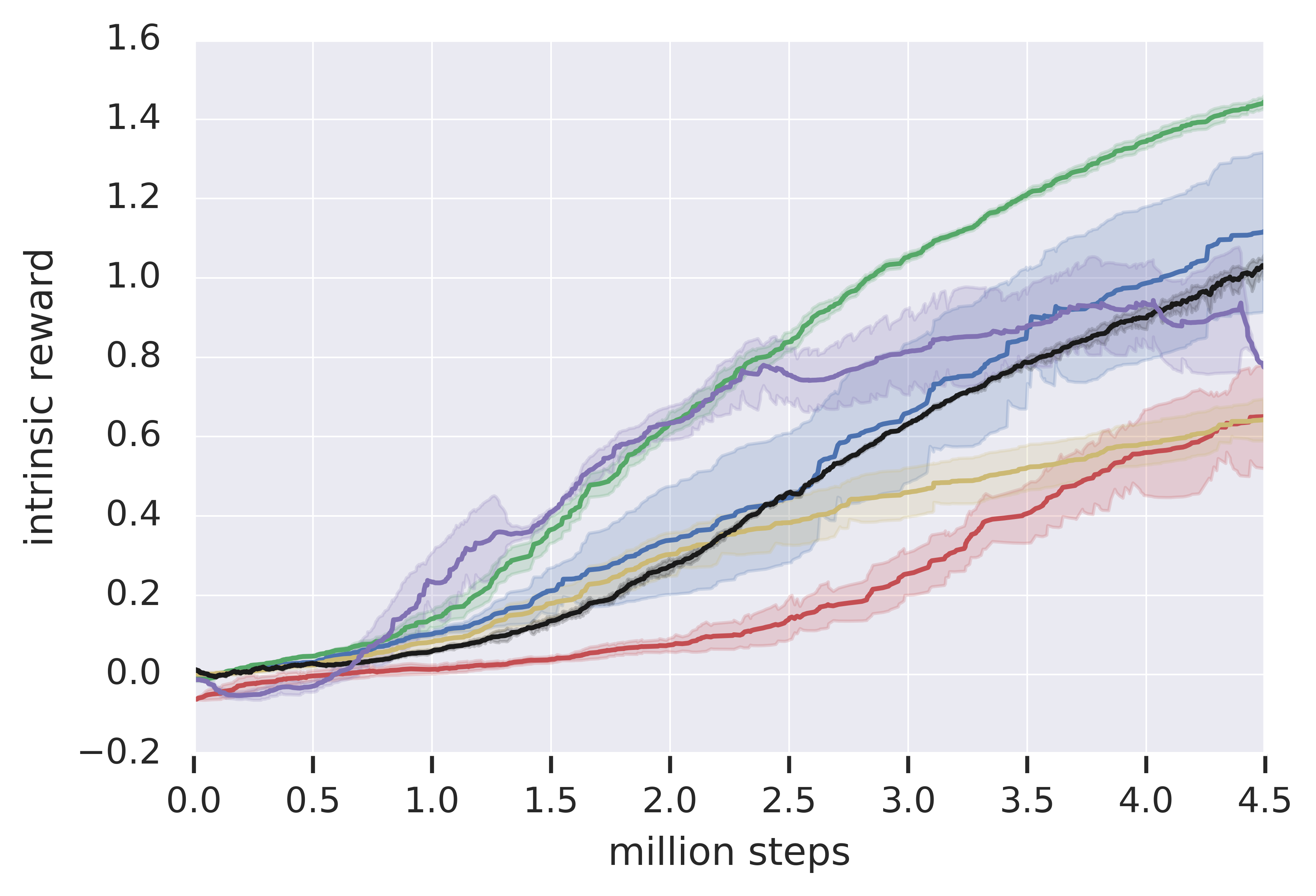}
    \end{minipage}
    \begin{minipage}{0.24\textwidth}\centering
    \small (c) Ant (x-y)\par 
    \includegraphics[width=\textwidth]{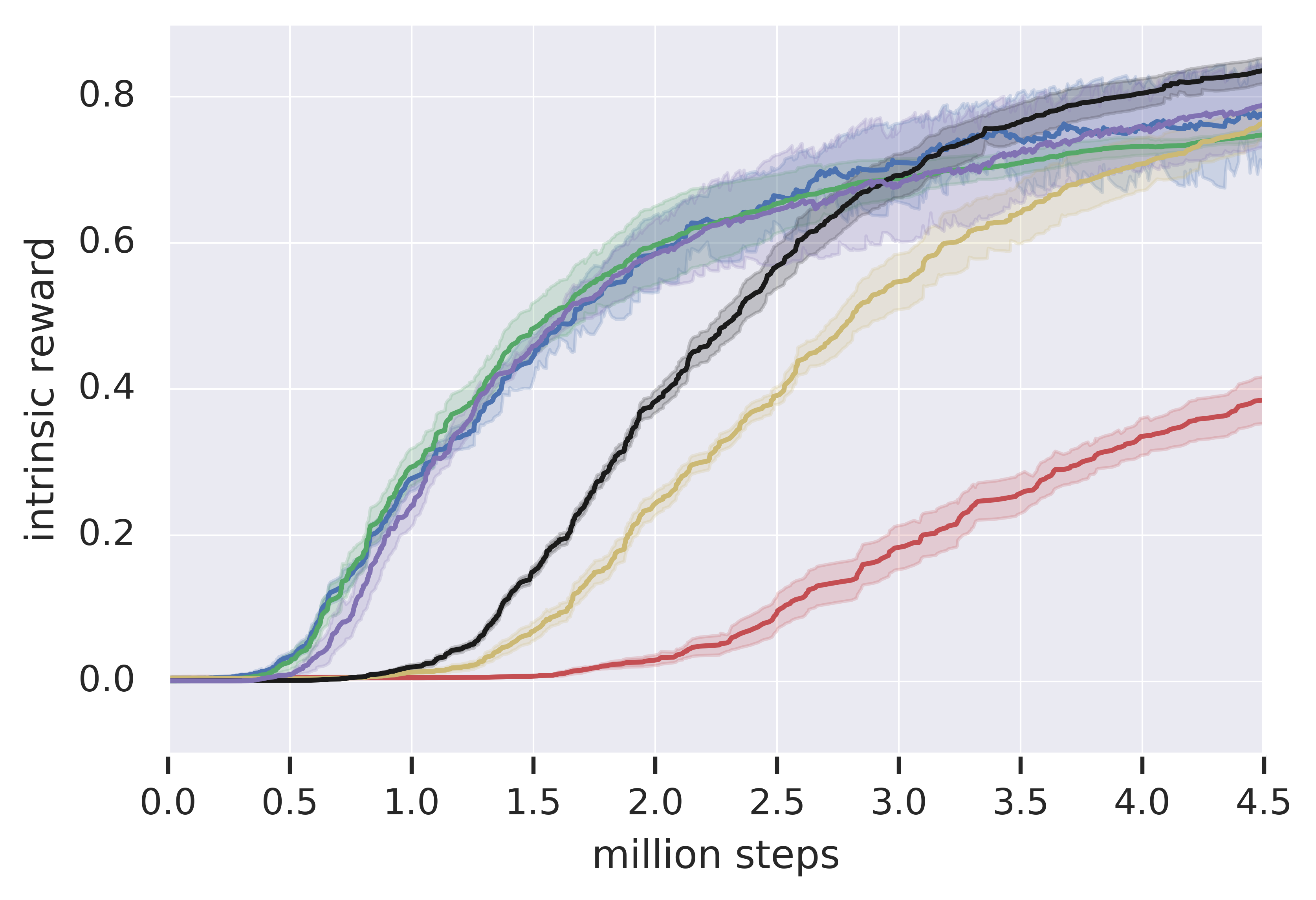}
    \small (d) Humanoid\par
    \includegraphics[width=\textwidth]{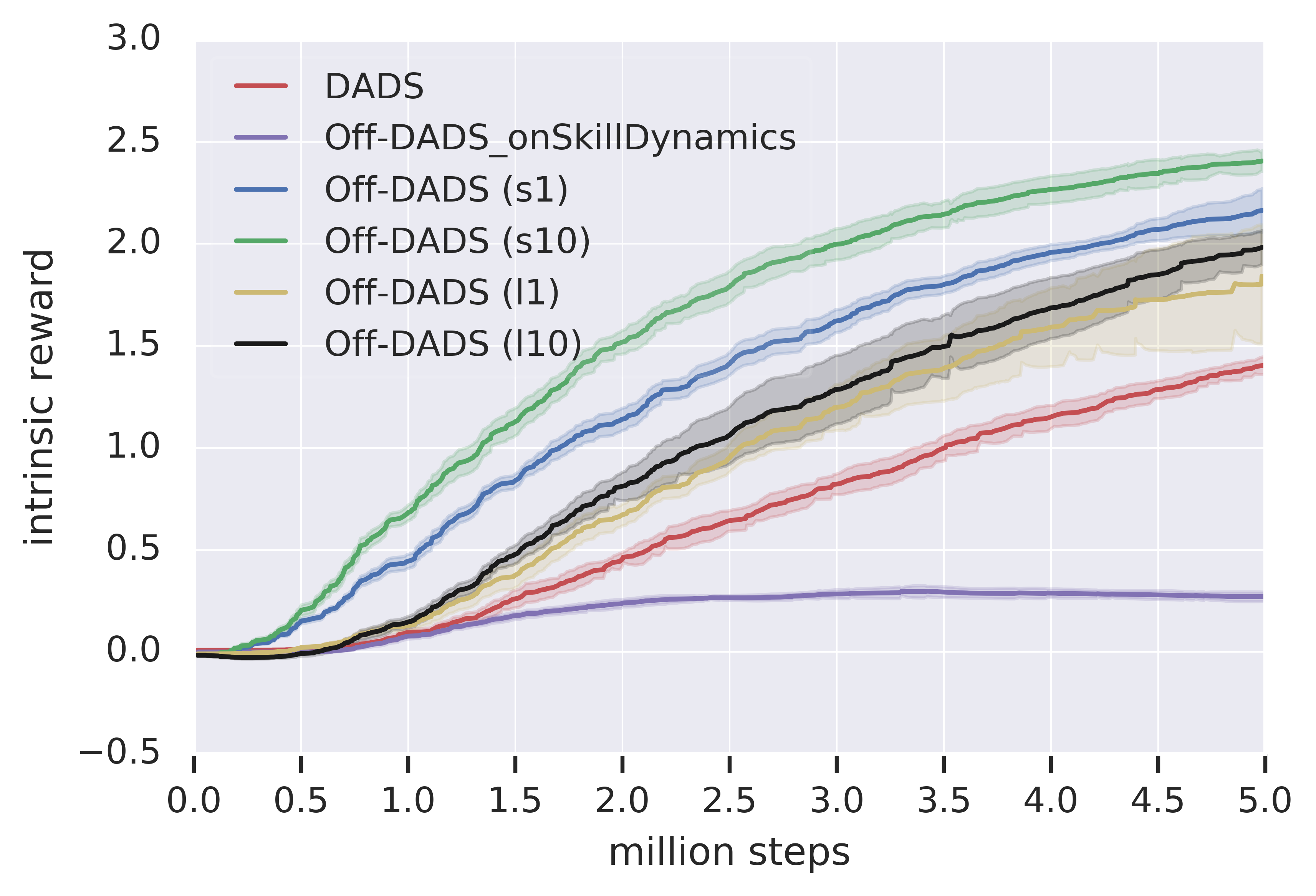}
    \end{minipage}
    \caption{\small Sample efficiency comparison of off-DADS with DADS (red). We control the effect of state-distribution shift using length of replay buffers (s implies short and l implies long replay buffer) and importance sampling corrections (1 and 10 being the values of the clipping parameter). We observe that all variants of off-DADS outperforms DADS in terms of sample efficiency, and using a short replay buffer with importance sampling clip parameter set to 10 consistently gives the best performance.}
    \label{fig:reward_curves}
\end{figure}
We benchmark off-DADS and its variants on continuous control environments from OpenAI gym \cite{1606.01540}, similar to \cite{sharma2019dynamics}. We use the HalfCheetah, Ant, and Humanoid environments, with state-dimensionality 18, 29, and 47 respectively. We also consider the setting where the skill-dynamics only observes the global $x, y$ coordinates of the Ant. This encourages the agent to discover skills which diversify in the $x,y$ space, yielding skills which are more suited for locomotion \citep{eysenbach2018diversity, sharma2019dynamics}. 

To evaluate the performance of off-DADS and the role of hyperparameters, we consider the following variantions:
\begin{itemize}
    \item \textit{Replay Buffer Size}: We consider two sizes for the replay buffer $\mathcal{R}$: 10,000 (s) and 1,000,000 (l). As alluded to, this controls how \textit{on-policy} the algorithm is. A smaller replay buffer will have lower bias due to state-distribution shift, but can lose sample efficiency as it discards samples faster~\citep{gu2017interpolated}.
    \item \textit{Importance Sampling}: We consider two settings for the clipping parameter in the importance sampling correction: $\alpha=1$ and $\alpha=10$. The former implies that there is no correction as all the weights are clipped to 1. This helps evaluate whether the suggested importance sampling correction gives any gains in terms of sample efficiency.
\end{itemize}
This gives us four variants abbreviated as \textit{s1}, \textit{s10}, \textit{l1} and \textit{l10}. We also evaluate against the off-DADS variant where the skill-dynamics is trained on on-policy samples from the current policy. This helps us evaluate whether training skill-dynamics on off-policy data can benefit the sample efficiency of off-DADS. Note, while this ablation helps us understand the algorithm, this scheme would be wasteful of samples in asynchronous off-policy real world training, where the data from different actors could potentially be coming from different (older) policies. Finally, we benchmark against the baseline DADS, as formulated in \citep{sharma2019dynamics}. The exact hyperparameters for each of the variants are listed in Appendix~\ref{app:hyperparameters}. We record curves for five random seeds for the average intrinsic reward $r(s, z, s')$ as a function of samples from the environment and report the average curves in Figure~\ref{fig:reward_curves}.

We observe that all variants of off-DADS consistently outperform the on-policy baseline DADS on all the environments. The gain in sample efficiency can be as high as four times, as is the case for Ant (x-y) environment where DADS takes 16 million samples to converge to the same levels as shown for off-DADS (about 0.8 average intrinsic reward). We also note that irrespective of the size of the replay buffer, the importance sampling correction with $\alpha=10$ outperforms or matches $\alpha=1$ on all environments. This positively indicates that the devised importance sampling correction makes a better bias-variance trade-off than no importance sampling. The best performing variant on every environment except Ant (x-y) is the \textit{s10}. While training skill-dynamics on-policy provides a competitive baseline, the short replay buffer and the clipped importance sampling counteract the distribution shift enough to benefit the overall sample efficiency of the algorithm. Interestingly on Ant (x-y), the best performing variant is \textit{l10}. The long replay buffer variants are slower than the short replay buffer variants but reach a higher average intrinsic reward. This can be attributed to the smaller state-space for skill-dynamics (only $2$-dimensional) and thus, the state-distribution correction required is potentially negligible but at the same time the off-policy data is helping learn better policies.

\subsection{Real-world Training}
\label{sec:experimental_details}
\begin{figure}[!htb]
    \centering
    \begin{minipage}{0.48\textwidth}
    \centering
    \includegraphics[width=0.45\textwidth]{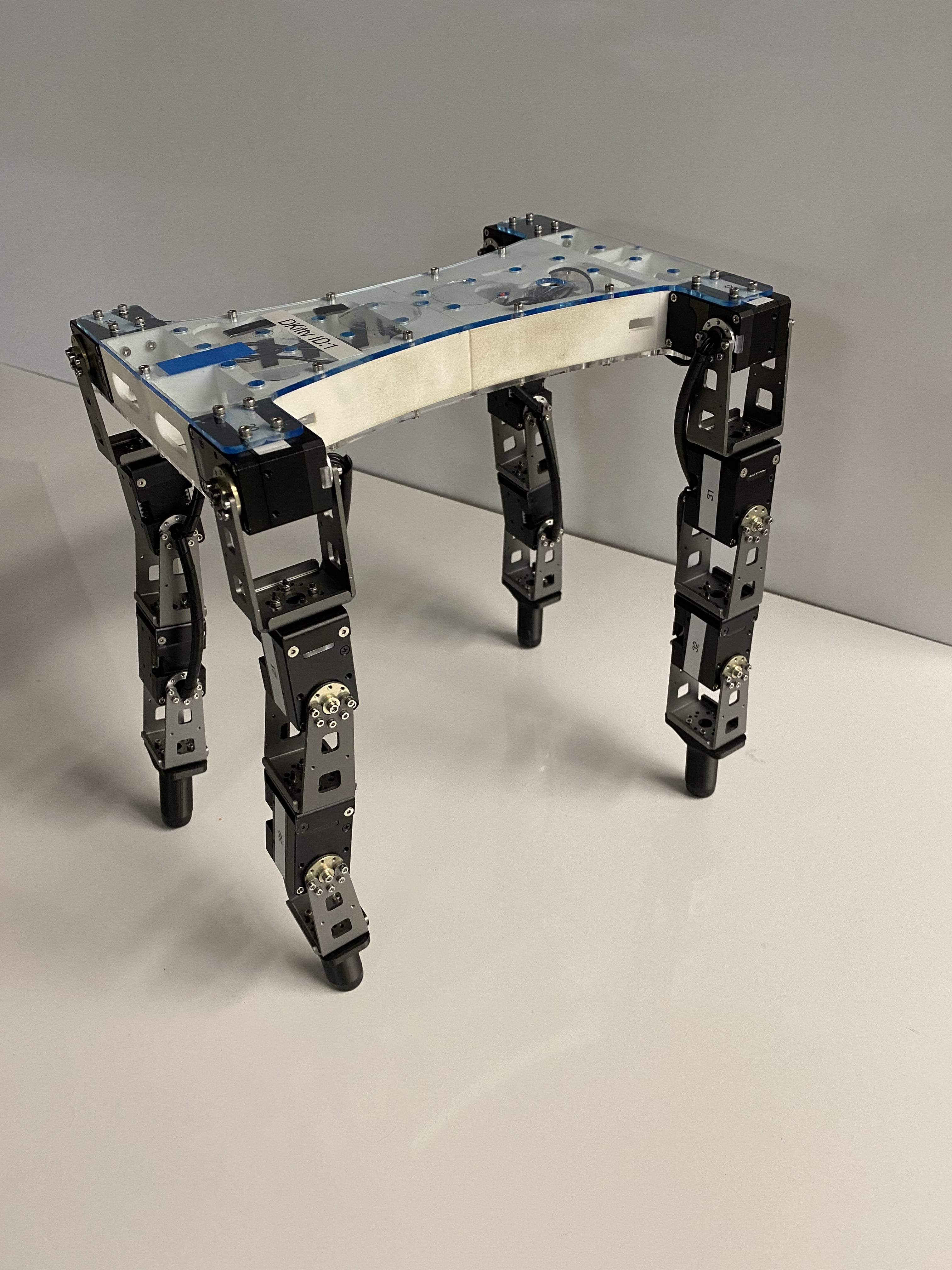}
    \includegraphics[width=0.45\textwidth]{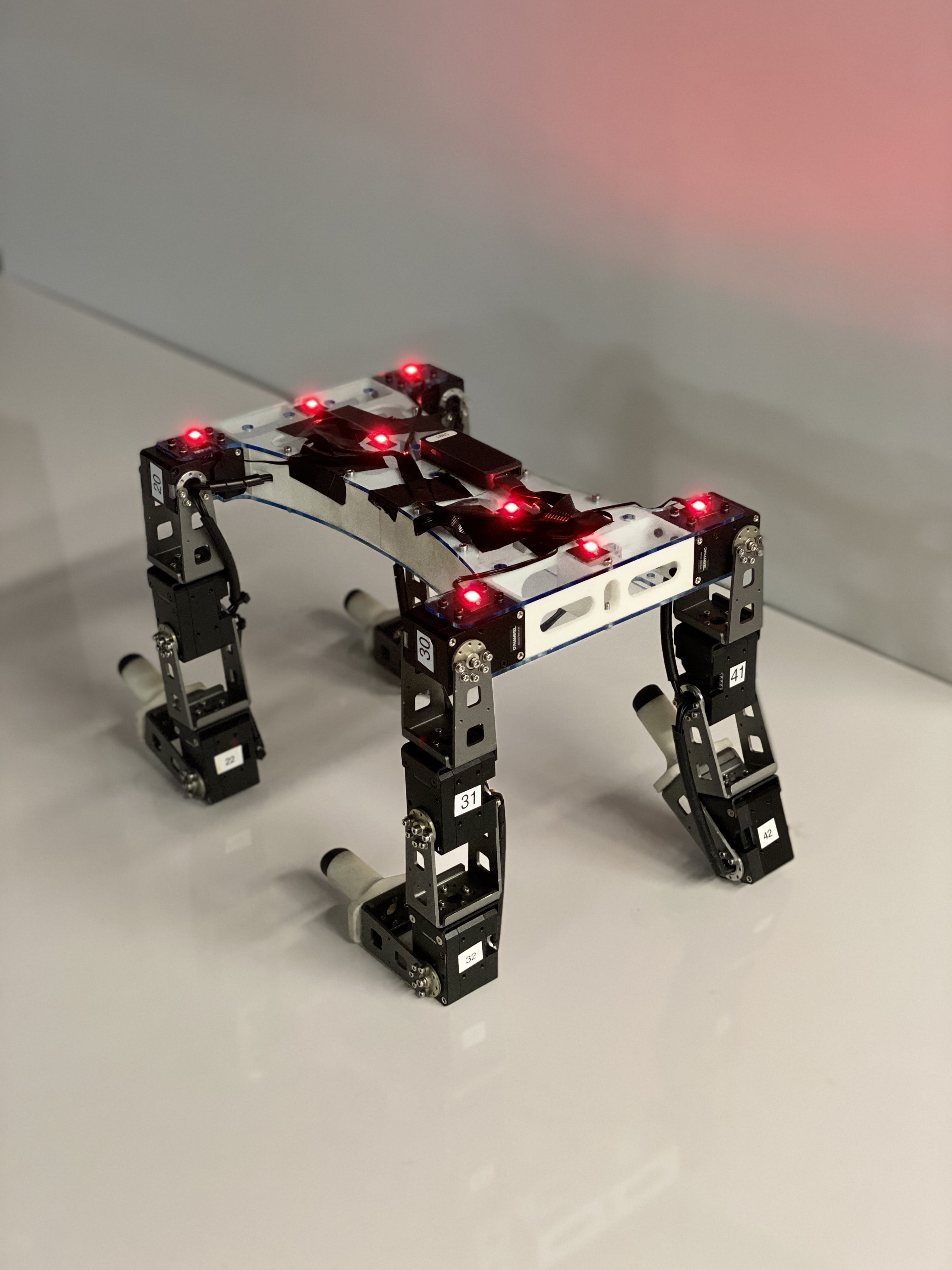}
    \end{minipage}
    \caption{\small (Left) D'Kitty robot from the ROBEL benchmark. (Right) D'Kitty with the LED configuration for PhaseSpace tracking.}
    \label{fig:dkitty_hardware}

\end{figure}

We now demonstrate the off-DADS can be deployed for real world reward-free reinforcement learning. To this end, we choose the ROBEL benchmark \citep{ahn2019robel}. In particular, we deploy off-DADS on D'Kitty shown in the Figure~\ref{fig:dkitty_hardware}. D'Kitty is a 12 DOF compact quadruped capable of executing diverse gaits. We also provide preliminary results for D'Claw, a manipulation-oriented setup from ROBEL in Appendix~\ref{appendix: D'Claw}.

\subsection{D'Kitty Experimental Setup}
\begin{figure}[!htb]
    \centering
    \begin{minipage}{0.48\textwidth}
    \includegraphics[width=\textwidth]{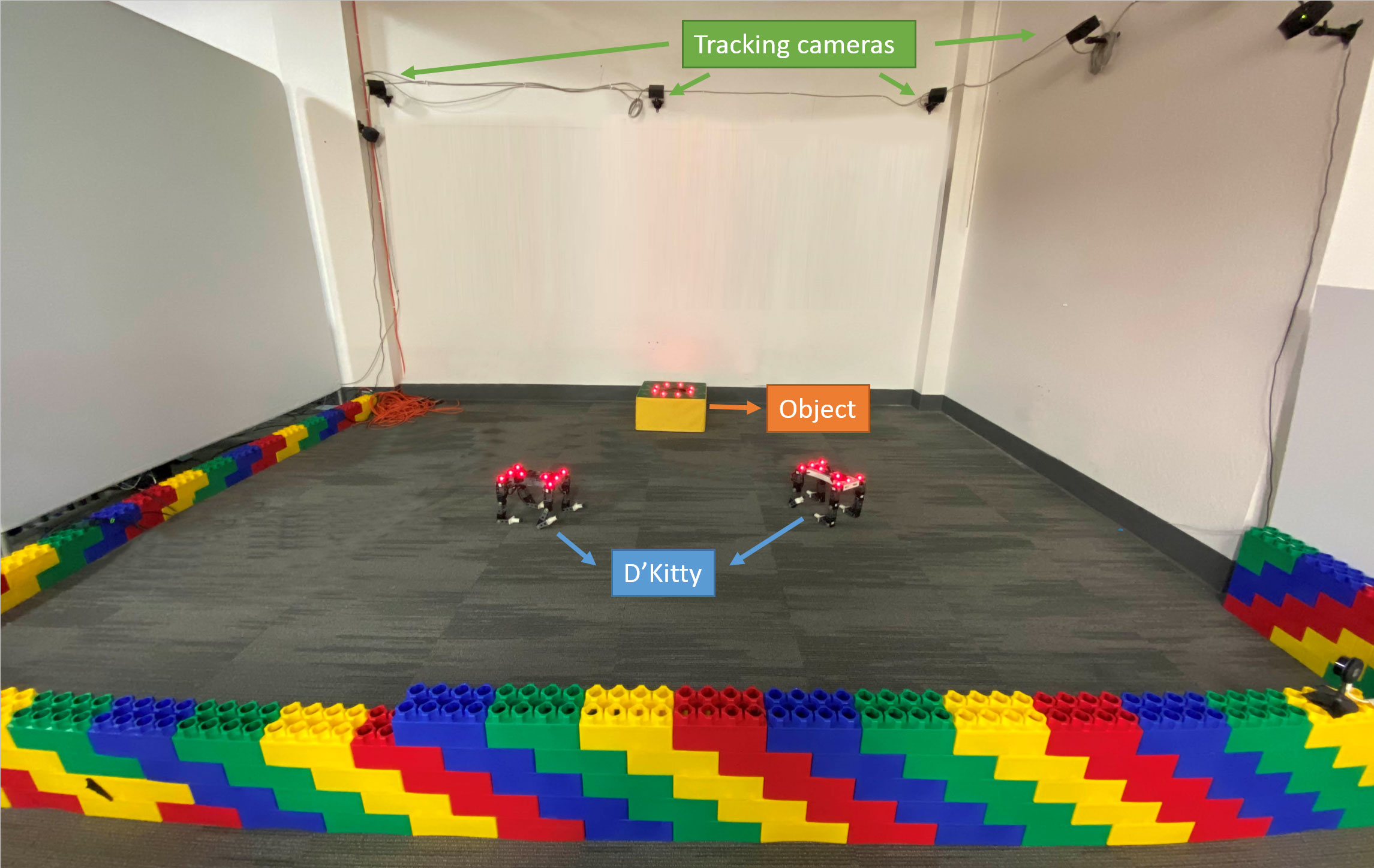}
    \end{minipage}
    \caption{\small Two quadrupeds in a cordoned area. The LEDs allow the robots and the object to be tracked using the PhaseSpace cameras.}
    \label{fig:play_area}
\end{figure}
To run real-world training, we constructed a walled $4m \times 4m$  cordoned area, shown in \autoref{fig:play_area}. The area is equipped with $5$ PhaseSpace Impulse X2 cameras that are equidistantly mounted along two bounding orthogonal walls. These cameras are connected to a PhaseSpace Impulse X2E motion capture system which performs 6 DOF rigid body tracking of the D'Kitty robots' chassis at 480Hz. 
We use two D'Kitty robots for data collection and training in experiment. Each D'Kitty, we attach one PhaseSpace LED controller which controls 8 active LED markers that are attached to the top surface of the D'Kitty chassis as shown in \autoref{fig:dkitty_hardware}. Each D'Kitty is tethered via 3 cables: USB serial to the computer running off-DADS, 12V/20A power to the D'Kitty robot, and USB power to the LED controller. To reduce wire entanglement with the robot, we also have an overhead support for the tethering wires.

\begin{figure}[!htb]
    \centering
    \begin{tabular}{|c|c|}
    \hline
    \textbf{Dynamixel Property} & \textbf{Value} \\
    \hline
    Model           & XM-W210 \\
    Control Mode    & Position Control \\
    Baudrate        & 1 Mbps           \\
    PWM Limit       & 450 (50.85\%)    \\
    Voltage Range   & 9.5V to 16V      \\
    \hline
    \end{tabular}
    \caption{\small Dynamixel motor configuration for the D'Kitty robots.}
    \label{fig:dynamixel_config}
\end{figure}

\subsection{Algorithmic Details}
We first test the off-DADS algorithm variants in simulation. For the D'Kitty observation space, we use the Cartesian position and Euler orientation (3 + 3), joint angles and velocities (12 + 12), the last action executed by the D'Kitty (12) and the upright (1), which is the cosine of the orientation with global $z$-axis. The concatenated observation space is 43-dimensional. Hyperparameter details for off-DADS (common to all variants) are as follows: The skill space $\mathcal{Z}$ is $2$D with support over $[-1,1 ]^2$. We use a uniform prior $p(z)$ over $\mathcal{Z}$. We parameterize $\pi(a \mid s, z), Q^{\pi}(s, a, z)$ and $q_\phi(s' \mid s, z)$ using neural networks with two hidden layers of size $512$. The output of $\pi(a \mid s, z)$ is parameterized by a normal distribution $\mathcal{N}(\mu, \Sigma)$ with a diagonal covariance which is scaled to $[-1, 1]$ using $tanh$ transformation. For $q_\phi$, we reduce the observation space to the D'Kitty co-ordinates $(x, y)$. This encourages skill-discovery for locomotion behaviors \citep{sharma2019dynamics, eysenbach2018diversity}. We parameterize $q_\phi$ to predict $\Delta s = s' - s$, a general trick in model-based control which does not cause any loss in representational power as the next state can be recovered by adding the prediction to the current state. We use soft-actor critic \citep{haarnoja2018soft} to optimize $\pi, Q^\pi$. To learn $q_\phi$, we sample batches of size $256$ and use the Adam optimizer with a fixed learning rate of $0.0003$ for $T_q = 8$ steps. For soft-actor critic, we again use Adam optimizer with a fixed learning rate of $0.0003$ while sampling batches of size $256$ for $128$ steps. Discount factor $\gamma=0.99$, with a fixed entropy coefficient of $0.1$. For computing the DADS reward $r(s ,z, s')$, we set $L=100$ samples from the prior $p(z)$. We set the episode length to be $200$, which terminates prematurely if the upright coefficient falls below $0.9$ (that is the D'Kitty is tilting more than 25 degrees from the global $z$-axis).

\begin{figure}[!htb]
    \centering
    \begin{minipage}{0.48\textwidth}
    \centering
    \includegraphics[width=\textwidth]{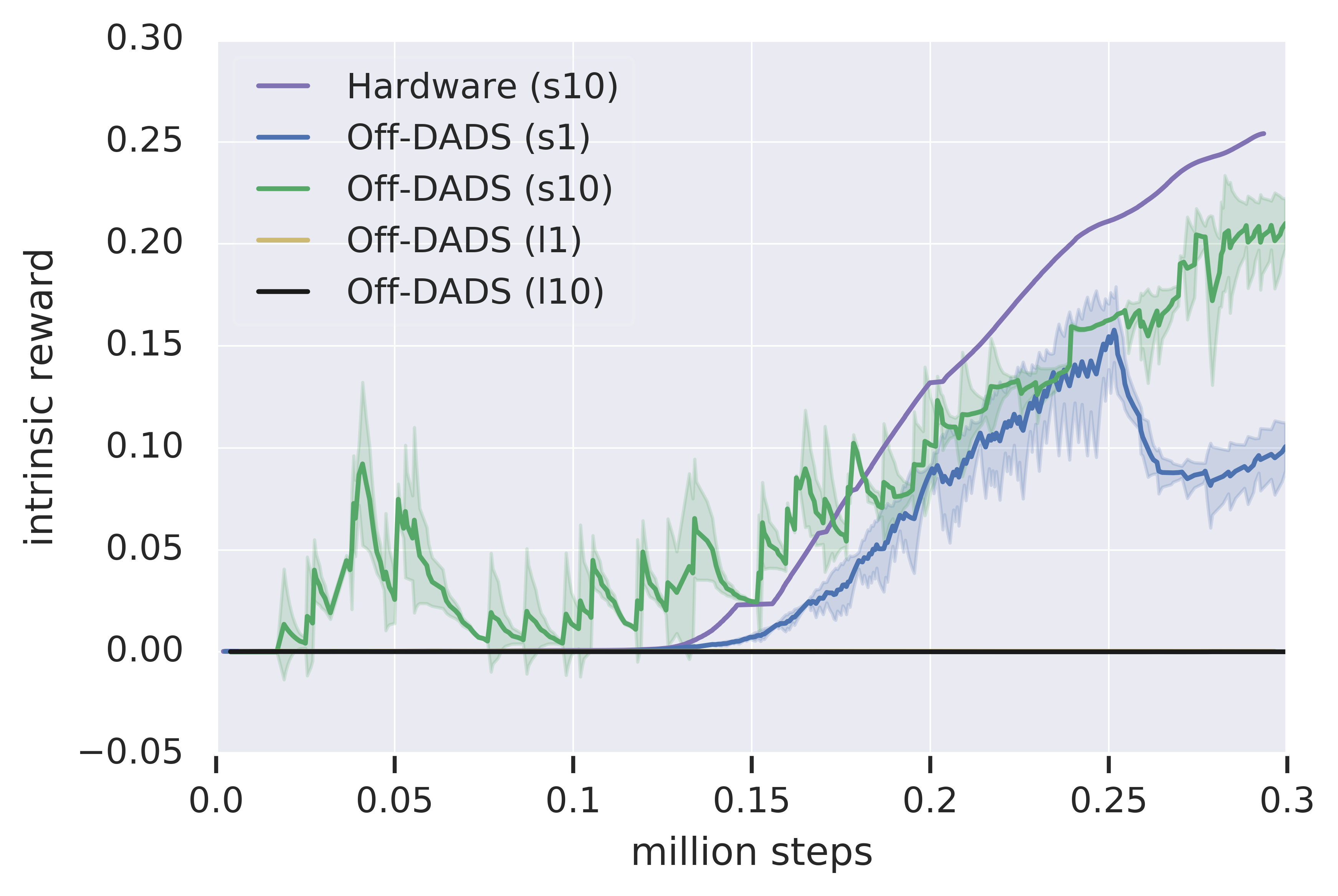}
    \end{minipage}
    \caption{\small (Left) Training curves for D'Kitty in both simulation and real-world. We find the off-DADS with a short replay and importance sampling clipping parameter $\alpha=10$ to be the most suitable for the real-world learning. We find the real-world learning curve closely follows the simulation learning curve.}
    \label{fig:dkitty_learning}
\end{figure}

In terms of off-DADS variants, we evaluate the four variants discussed in the previous section. For all the variants, we collect at least $500$ steps in the simulation before updating $q_\phi$ and $\pi$. The observations for the variants resemble those of the Ant (x-y) environment. We observe that the variants with a replay buffer of size $10,000$ are much faster to learn than the replay buffer of size $1,000,000$. Asymptotically, we observe the long replay buffer outperforms the short replay buffer though. We also observe setting $\alpha=10$ benefits the cause of sample efficiency.

For the real robotic experiment, we choose the hyperparameters $\mathcal{R}$ to be of size $10,000$ and we set $\alpha=10$. While asymptotically better performance is nice, we prioritized sample efficiency. For the real experiment, we slightly modify the collection condition. For every update of $q_\phi$ and $\pi$, we ensure there are $200$ new steps and at least $3$ new episodes in the replay buffer $\mathcal{R}$.

% Real DKitty v/s Sim D'Kitty
% Training setup, reset, episode filtering etc
% Skills in simulation, skills in real comparison

\subsection{Emergent Locomotion Behaviors}
\begin{figure}[!htb]
    \centering
    \begin{minipage}{0.48\textwidth}
    \includegraphics[width=\textwidth]{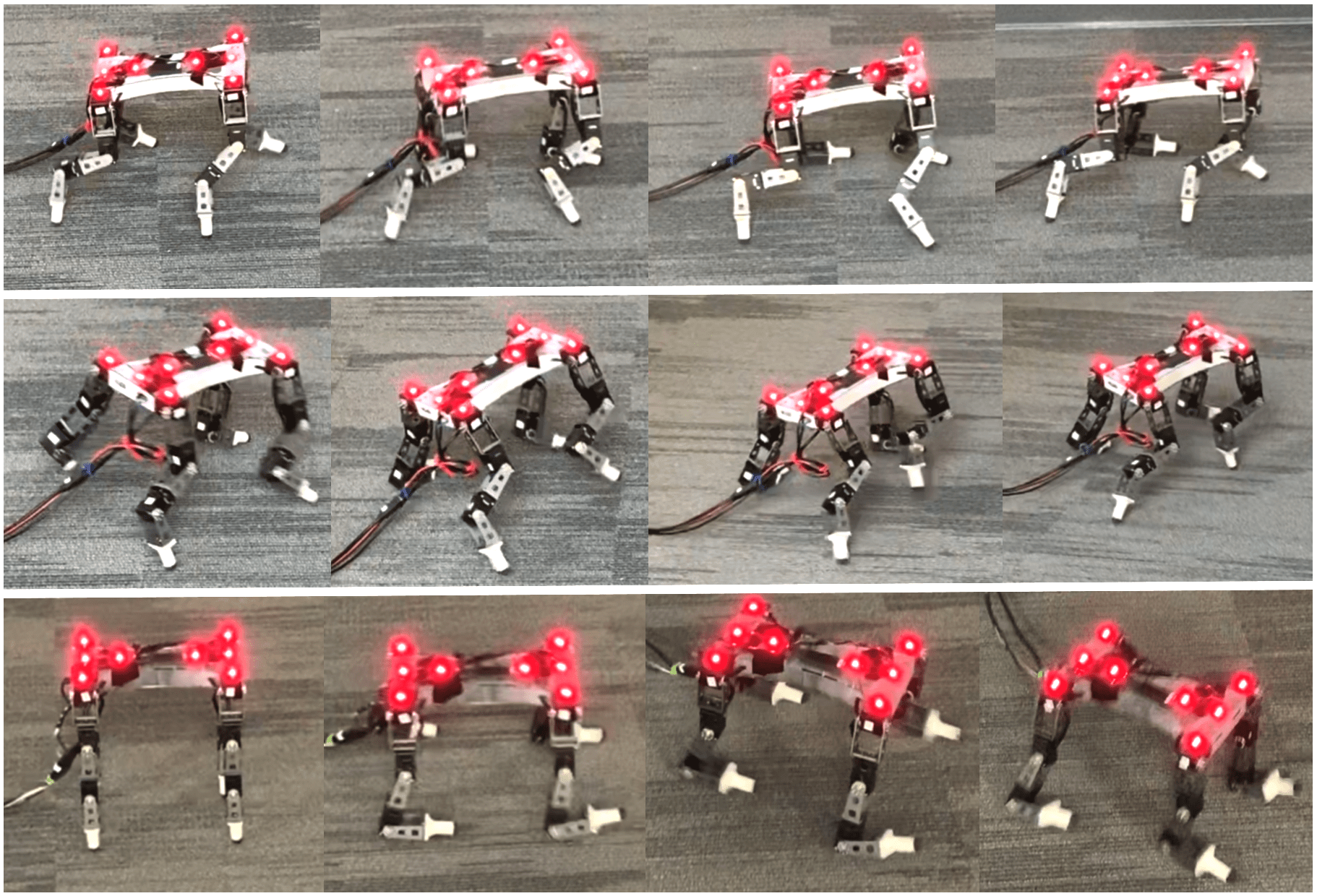}
    \end{minipage}
    \caption{\small Diverse gaits learned by the D'Kitty in our real world experiments.}
    \label{fig:dkitty_gaits}
\end{figure}
With the setup and hyperparameters described in the previous sections, we run the real-world experiment. The experiment was ran over 3 days, with the effective training time on the robot being 20 hours (including time spent in maintaining the hardware). We collected around $300,000$ samples in total as shown in the learning curve in \autoref{fig:dkitty_learning}. We capture the emergence of locomotion skills in our video supplement. \autoref{fig:superposed_skills} and \autoref{fig:dkitty_gaits} show some of the diversity which emerges in skills learned by D'Kitty using off-DADS, in terms of orientation and gaits.
% Full detail of the set-up, the learning progress, the resets, and the diversity of behaviors learned are available at following anonymous project's website: \href{https://sites.google.com/view/off-DADS}{https://sites.google.com/view/off-DADS}.

Broadly, the learning occurs in the following steps: (a) D'Kitty first tries to learn how to stay upright to prolong the length of the episode. This happens within the first hour of the episode. (b) It spends the next few hours trying to move around while trying to stay upright. These few hours, the movements are most random and the intrinsic reward is relatively low as they do not correlate well with $z$. (c) About 5-6 hours into the training, it starts showing a systematic gait which it uses to move in relatively random directions. This is when the intrinsic reward starts to rise. (d) A few more hours of training and this gait is exploited to predictably move in specific directions. At this point the reward starts rising rapidly as it starts diversifying the directions the agent can move in predictably. Interestingly, D'Kitty uses two different gaits to capture and grow in two different directions of motion, which can be seen in the video supplement. (e) At about 16 hours of training, it can reliably move in different directions and it is trying to further increase the directions it can move in predictably. Supplementary videos are available here: \href{https://sites.google.com/view/dads-skill}{https://sites.google.com/view/dads-skill}

One interesting difference from simulation where the D'Kitty is unconstrained, is that the harnesses and tethering despite best attempts restrain the movement of the real robot to some extent. This encourages the agent to invest in multiple gaits and use simpler, more reliable motions to move in different directions.
\subsection{Challenges in real-world training}
\label{sec:challenges}
We discuss some of the challenges encountered during real-world reinforcement learning, particularly in context of locomotive agents. 
\begin{itemize}
    \item Reset \& Autonomous operation: A good initial state distribution is necessary for the exploration to proceed towards the desirable state distribution. In context of locomotion, a good reset comprises of being in an upright position and relocating away from the extremities of the area. For the former, we tried two reset mechanisms: (a) scripted mechanism, which is shown in the supplementary video and (b) reset detector which would continue training if the D'Kitty was upright (based on height and tilt with z-axis), else would wait (for human to reset). However, (a) being programmatic is not robust and does not necessarily succeed in every configuration, in addition to being slow. (b) can be really fast considering that D'Kitty is reasonably compact, but requires human oversight. Despite human oversight, the reset detector can falsely assume the reset is complete and initiate the episode, which requires an episode filter to be written.
    Relocating from the extremities back to the center is a harder challenge. It is important because the tracker becomes noisy in those regions while also curbing the exploration of the policy. However, this problem only arises when the agent shows significant skills to navigate. There are other challenges besides reset which mandate human oversight into the operation. Primarily, random exploration can be tough on the robot, requiring maintenance in terms of tightening of screws and sometimes, replacing motors. We found latter can be avoided by keeping the motor PWMs low (450 is good). While we make progress towards reward-free learning in this work, we leave it to future work to resolve problems on the way to fully autonomous learning.
    \item Constrained space: For the full diversity of skills to emerge, an ideal operation would have unconstrained space with accurate tracking. However, realistically that is infeasible. Moreover, real operation adds unmodelled constraints over simulation environments. For example, the use of harness to reduce wire entanglement with the robot adds tension depending on the location. When operating multiple D'Kitties in the same space, there can be collisions during the operation. Likelihood of such an event progressively grows through the training. About two-fifths through the training, we started collecting with only one D'Kitty in the space to avoid future collisions. Halfway through the training, we decided to expand the area to its limits and re-calibrate our tracking system for the larger area. Despite the expansion, we were still short on space for the operation of just one D'Kitty. To remedy the situation, we started decreasing the episode length. We went from 200 steps to 100, then to 80 and then to 60. However, we observed that short episodes started affecting the training curve adversarially (at about 300k samples). Constrained by the physical limits, we finished the training. While a reasonable skill diversity already emerges in terms of gaits and orientations within the training we conduct, as shown in \autoref{fig:superposed_skills}, more skills should be discovered with more training (as suggested by the simulation results as well as the fact the reward curve has not converged). Nonetheless, we made progress towards the conveying the larger point of reward-free learning being realized in real-world.
    
\end{itemize}

\subsection{Harnessing Learned Skills}
\begin{figure}[!htb]
    \centering
    \begin{tabular}{|c|c|}
    \hline
    Distance Travelled & $2.17 \pm 0.59$ \\
    \hline
    Percentage of Falls   & 5\% \\
    \hline
    \end{tabular}
    \caption{\small Distance (in m) travelled by skills sampled randomly from the prior in 100 steps, which is 10s of execution time. We also report the number of times D'Kitty falls before completing 100 steps. The results have been averaged over 20 trials.}
    \label{table:traj_stats}
\end{figure}

\begin{figure}[!htb]
    \centering
    \includegraphics[width=0.48\textwidth]{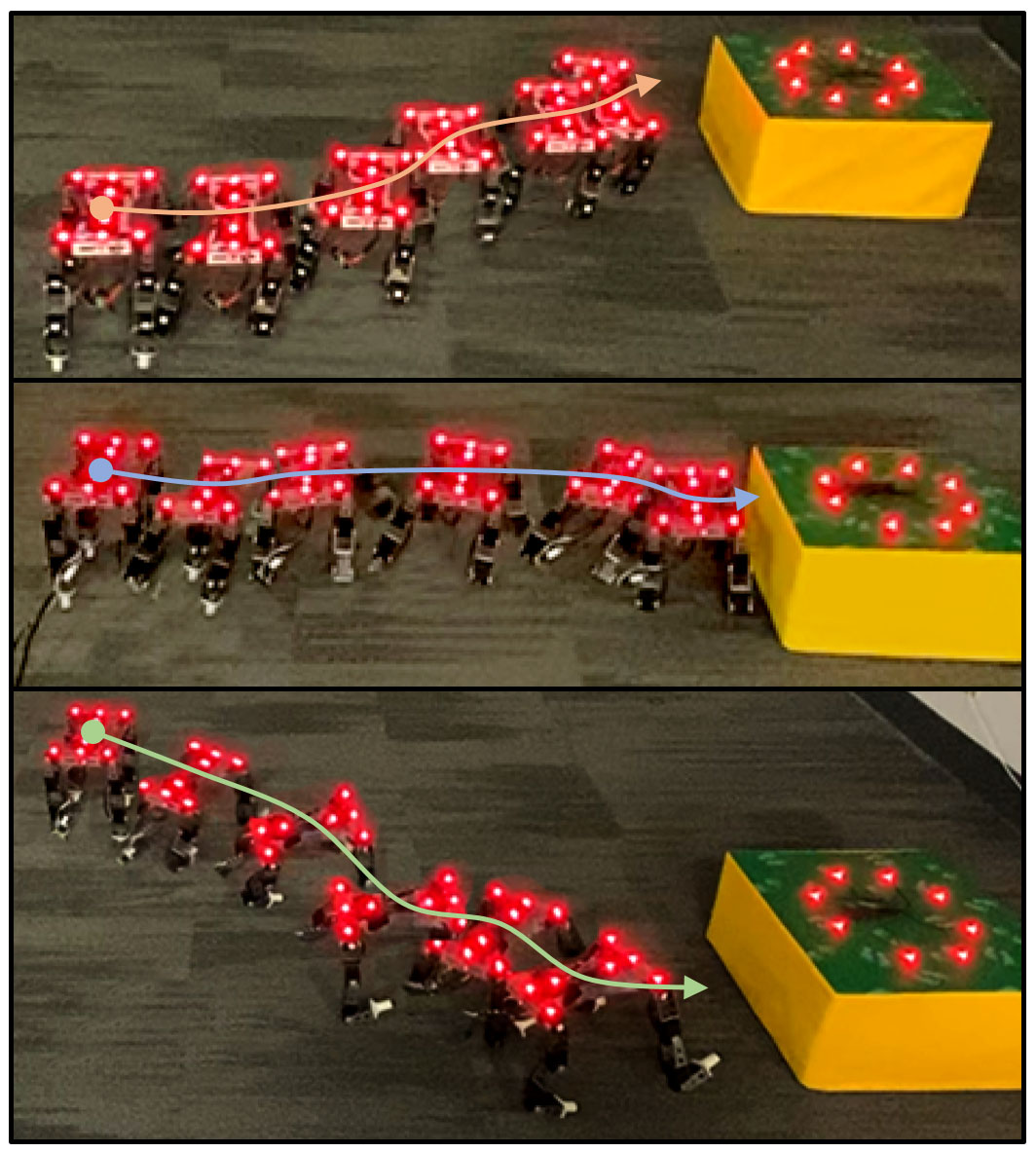}
    \caption{\small Navigation via model-predictive-control over skills learned by off-DADS. Using the skill-dynamics learned in the unsupervised training, the planner composes skills to move towards the goal. The trajectories visualized show movement of D'Kitty towards the goal box, marked by a set of LED markers.}
    \label{fig:goalnav_mpc}
\end{figure}
Qualitatively, we see that a diverse set of locomotive skills can emerge from reward-free training. However, as has been discussed in \citep{sharma2019dynamics}, these skills can be harnessed for downstream tasks using model-based control on the learned skill-dynamics $q_\phi(s'\mid s,z)$. First, we partly quantify the learned skills in \autoref{table:traj_stats}. We execute skills randomly sampled from the prior and collect statistics for these runs. In particular, we find that despite limited training, the skills are relatively robust and fall in only $5\%$ of the runs, despite being proficient in covering distance.   
Interestingly, the learned skills can also be harnessed for model-based control as shown in \autoref{fig:goalnav_mpc}. The details for model-predictive control follow directly from \citep{sharma2019dynamics}, which elucidates on how to to do so using skill-dynamics $q_\phi$ and $\pi$. We have included video supplements showing model-predictive control in the learned skill space for goal navigation.
\section{Conclusion}
\label{sec:conclusion}
In this work, we derived off-DADS, a novel off-policy variant to mutual-information-based reward-free reinforcement learning framework. The improved sample-efficiency from off-policy learning enabled the algorithm to be applied on a real hardware, a quadruped with 12 DoFs, to learn various locomotion gaits under 20 hours without human-design reward functions or hard-coded primitives. Given our dynamics-based formulation from~\citep{sharma2019dynamics}, we further demonstrate those acquired skills are directly useful for solving downstream tasks such as navigation using online planning with no further learning. We detail the successes and challenges encountered in our experiments, and hope our work could offer an important foundation toward the goal of unsupervised, continual reinforcement learning of robots in the real world for many days with zero human intervention.

% \section*{Acknowledgments}

\bibliographystyle{plainnat}
\bibliography{references}

\newpage

\appendix
\subsection{Expanding the stationary distribution}
\label{app:stationary}
\noindent Recall that the stationary distribution can be written as $$p(s \mid z) = \sum_{t=0}^{T} \gamma^{t} p(s_t = s \mid z)$$
for a $\gamma$-discounted episodic setting of horizon $T$. Here, 
$$p(s_t \mid z) = \int p_0(s_0) \prod_{i=0}^{t-1} \pi(a_i \mid s_i, z) p(s_{i+1} \mid s_i, z) da_i ds_i$$
Now, using this in the definition of $J(\pi)$, we get
\begin{multline*}
    J(\pi) = \int p(z) \Big[\sum_{t=0}^{T-1}\gamma^t \int p_0(s_0) \prod_{i=0}^{t-1} \pi(a_i \mid s_i) \\
    p(s_{i+1} \mid s_i, a_i) ds_i da_i \Big] p(s' \mid s, z) r(s, z, s')dz ds ds'
\end{multline*}
Now, we can switch the integral for variables $s, s'$ and the summation over index $t$, distributing the terms $p(s' \mid s, z) r(s, z, s')$ over $T$ in the summation. For each term, we can rename the variable $s \rightarrow s_t$ and $s' \rightarrow s_{t+1}$. We get,
\begin{multline*}
    J(\pi) = \int p(z) \Big[\sum_{t=0}^{T-1} \int p_0(s_0) \prod_{i=0}^{t} \pi(a_i \mid s_i) p(s_{i+1} \mid s_i, a_i) \\  \gamma^{t} r(s_t, z, s_{t+1})ds_i da_i \Big] dz
\end{multline*}
where we have also expanded $p(s_{t+1} \mid s_t, z) = \int p(s_{t+1} \mid s_t, a_t) \pi(a_t \mid s_t, z) da_t$. Now, consider
\begin{multline*}
     \Big(\int p_0(s_0) \prod_{i=0}^{t} \pi(a_i \mid s_i) p(s_{i+1} \mid s_i, a_i) ds_i da_i\Big)\\\Big(\int \prod_{i=t+1}^{T-1}\pi(a_i \mid s_i) p(s_{i+1} \mid s_i, a_i) ds_i da_i\Big) = \\\int p_0(s_0)\prod_{i=0}^{T-1}\pi(a_i \mid s_i) p(s_{i+1} \mid s_i, a_i) ds_i da_i
\end{multline*}
However, $\int \prod_{i=t+1}^{T-1}\pi(a_i \mid s_i) p(s_{i+1} \mid s_i, a_i) ds_i da_i = 1$, as integral of a probability distribution over its support integrates to one. Therefore by introducing the dummy variables for the leftover trajectory in each term, the probability distribution in every term of the sum becomes the same. Exchanging the sum and the integral again, we can rewrite $J(\pi)$ as the desired expectation over all possible trajectories:
\begin{multline*}
J(\pi) = \int p(z) p_0(s_0) \prod_{t=0}^{T-1} \pi(a_t \mid s_t) p(s_{t+1} \mid s_t, a_t) \\
\Big(\sum_{t=0}^{T-1}\gamma^t r(s_t, z, s_{t+1}) \Big) dz ds_0 da_0 \ldots ds_{T}
\end{multline*}
This same exercise can be repeated for $J(q_\phi)$.

\subsection{Importance sampling corrections for skill dynamics}
\label{app:importance_sampling}

\noindent To derive the unbiased off-policy update for $q_\phi$, consider the defintion again:
\begin{align*}
    J(q_\phi) &= \int p(z) p(s \mid z) p(s' \mid s, z) \log q_\phi(s' \mid s, z) dz ds ds'\\
              &= \mathbb{E}_{a \mid s \sim \pi}\big[\sum_{t=0}^{T-1}\gamma^t \log q_\phi(s_{t+1} \mid s_t, z)\big]
\end{align*}
This results from the expansion of the state distribution $p(s \mid z)$, as discussed in Appendix~\ref{app:stationary}. Note, the expectation is taken with respect to trajectories generated by $\pi(a \mid s, z)$ for $z \sim p(z)$, and is contracted for convenience. Assuming the samples are generated from a behavior policy $\pi_c(a \mid s, z)$, we can rewrite $J(q_\phi)$ as follows:
$$J(q_\phi) = \mathbb{E}_{a\mid s\sim \pi_c}\Big[\sum_{t=0}^{T-1}\gamma^t \Big(\prod_{i=0}^t\frac{\pi(a_i \mid s_i, z)}{\pi_c(a_i \mid s_i, z)}\Big)\log q_\phi(s_{t+1} \mid s_t, z)\Big]$$
This results from the fact that the state at time $t$ only depends on actions preceding time $t$. Note, while this estimator is unbiased, it can be extremely unstable computationally as it is a product of probabilities which can become diminishingly small.

\subsection{Hyperparameters}
\label{app:hyperparameters}
The hyperparameters for all experiments heavily borrow from those in \autoref{sec:experimental_details}.

\subsubsection{HalfCheetah}
We learn a $3$D skill space with a uniform prior $[-1, 1]^2$. For on-policy DADS, we collect a batch of $2000$ steps in every iteration. The $\pi, Q^\pi$ is trained for 64 steps with a batch size of 256 sampled uniformly from the $2000$ steps. The off-policy version, has similar hyperparameters, except that $\pi$ samples from a replay buffer of size $100,000$, while the skill dynamics $q_\phi$ is only trained on the new batch of $2000$ steps. For all versions of off-DADS, we collect a fresh $1000$ samples in every iteration. Rest of the hyperparameters are the same.

\subsubsection{Ant}
For the Ant environment on the full state-space, we learn a $3$D skill space with a uniform prior $[-1, 1]^2$. For on-policy DADS, we collect a batch of $2000$ steps in every iteration. The $\pi, Q^\pi$ is trained for 64 steps with a batch size of 256 sampled uniformly from the $2000$ steps. The off-policy version, has similar hyperparameters, except that $\pi$ samples from a replay buffer of size $100,000$, while the skill dynamics $q_\phi$ is only trained on the new batch of $2000$ steps. For all versions of off-DADS, we collect a fresh $1000$ samples in every iteration. Rest of the hyperparameters are the same.

For the Ant environment on the $x$, $y$ space for the skill dynamics, we learn a $2$D skill space with a uniform prior $[-1, 1]^2$. For on-policy DADS, we collect a batch of $2000$ steps in every iteration. The $\pi, Q^\pi$ is trained for 64 steps with a batch size of 256 sampled uniformly from the $2000$ steps. The off-policy version, has similar hyperparameters, except that $\pi$ samples from a replay buffer of size $100,000$, while the skill dynamics $q_\phi$ is only trained on the new batch of $2000$ steps. For all versions of off-DADS, we collect a fresh $500$ samples in every iteration. Rest of the hyperparameters are the same.

\subsubsection{Humanoid}
We learn a $5$D space with a uniform prior $[-1, 1]^2$. For on-policy DADS, we collect a batch of $4000$ steps in every iteration. The $\pi, Q^\pi$ is trained for 64 steps with a batch size of 256 sampled uniformly from the $2000$ steps. The off-policy version, has similar hyperparameters, except that $\pi$ samples from a replay buffer of size $100,000$, while the skill dynamics $q_\phi$ is only trained on the new batch of $4000$ steps. For all versions of off-DADS, we collect a fresh $2000$ samples in every iterations Rest of the hyperparameters are the same.

\subsection{D'Claw (Simulation)}
\label{appendix: D'Claw}

\begin{figure}[!htb]
\centering
\begin{minipage}{0.24\textwidth}
    \centering
    \includegraphics[width=0.87\textwidth]{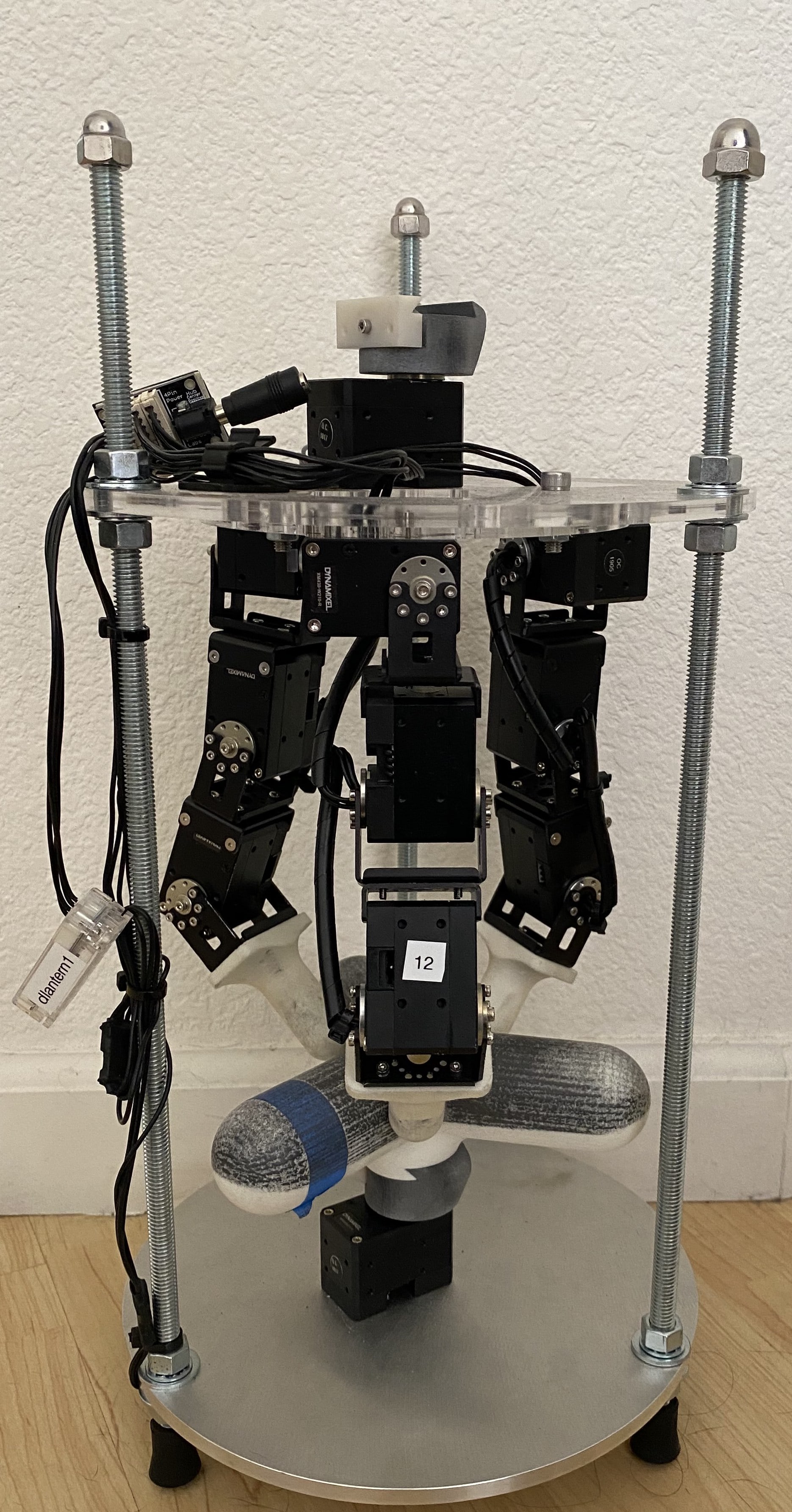}
\end{minipage}
\begin{minipage}{0.24\textwidth}
    \centering
    \includegraphics[width=\textwidth]{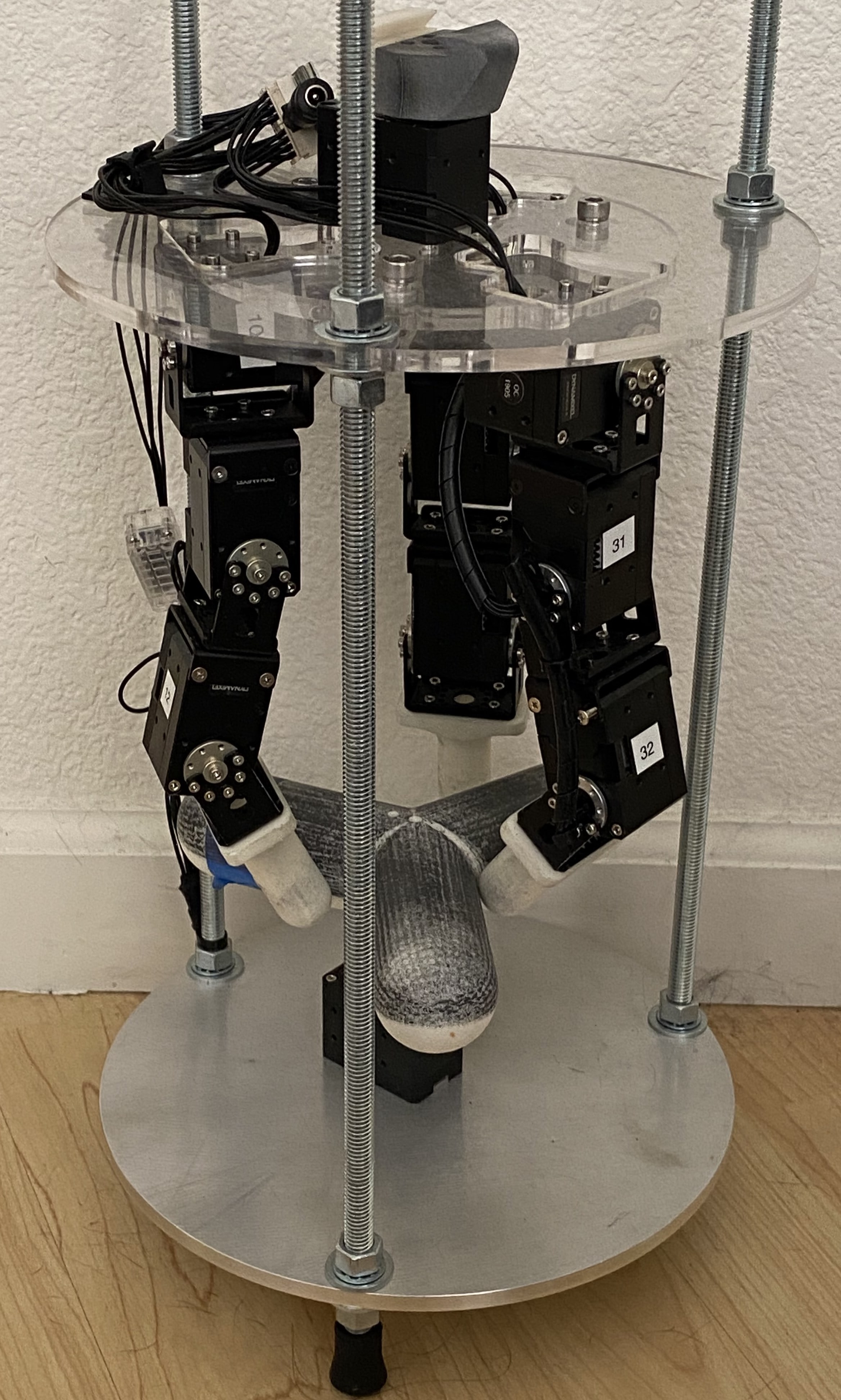}
\end{minipage}
\caption{D'Claw from ROBEL \citep{ahn2019robel} is a three-armed manipulatation oriented robotic setup. In this environment, it is provided with a three-pronged valve which can rotate on its axis. The goal position can be specified by the motor at the top.}
\label{fig:dclaw}
\end{figure}

We also provide additional results for the application of off-DADS to the manipulation domains. We choose the D'Claw environment from ROBEL, shown in Figure~\ref{fig:dclaw}. D'Claw is three independent arms with 9DOF capable of interacting and manipulating objects. For our setup, we have a three-pronged valve which can rotate about the z-axis. For the observation space in the environment, we use the euler angle of the valve (1), joint angles and velocities (9 + 9). The concatenated observation space is 19-dimensional. 

We provide results for the parallel simulation environment, which we will extend to hardware in the future. The hyperparameters details are as follows: We have a continuous $1D$ skill space $\mathcal{Z}$ with support over $[-1, 1]$. We use a uniform prior $p(z)$ over $\mathcal{Z}$. We parameterize $\pi(a \mid s, z), Q^\pi(s, a, z)$ and $q_\phi(s' \mid s, z)$ using neural networks with two hidden layers of size $512$. The output of $\pi(a \mid s, z)$ is parameterized by a normal distribution $\mathcal{N}(\mu, \Sigma)$ with a diagonal covariance matrix. The output of the normal distribution is scaled to $[-1, 1]$ using $tanh$ transformation. For $q_\phi$, we reduce the observation space to the valve orientation $(\theta)$. This encourages skill-discovery for behaviors which manipulate the object, and not just randomly move around their arms. We parameterize $q_\phi$ to predict $\Delta \theta = \theta' - \theta$. When predicting the next state using $\theta' = \theta + \Delta \theta$, we map the resulting orientation to $(-\pi, \pi]$. To learn $q_\phi$, we sample batches of size $256$ and use the Adam optimizer with a fixed learning rate of $0.0003$ for $T_q = 8$ steps. For soft-actor critic, we again use Adam optimizer with a fixed learning rate of $0.0003$ while sampling batches of size $256$ for $64$ steps. Discount factor $\gamma=0.99$, with a fixed entropy coefficient of $0.1$. For computing the DADS reward $r(s ,z, s')$, we set $L=100$ samples from the prior $p(z)$. We set the episode length to be $200$. We use a replay buffer $\mathcal{R}$ of size $100,000$ and set importance sampling clipping factor $\alpha=1$. 

\begin{figure}[!htb]
    \centering
    \subfigure{
        \centering
        \includegraphics[width=0.09\textwidth]{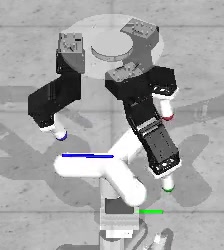}
        \includegraphics[width=0.09\textwidth]{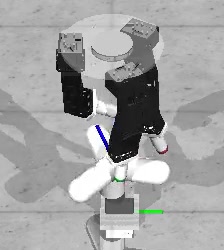}
        \includegraphics[width=0.09\textwidth]{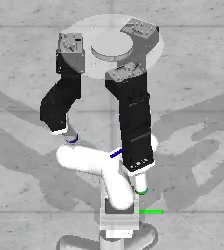}
        \includegraphics[width=0.09\textwidth]{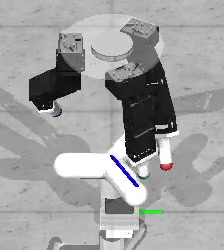}
        \includegraphics[width=0.09\textwidth]{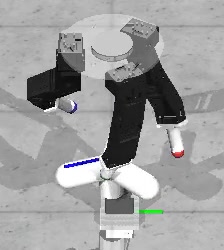}
    }
   \subfigure{
        \centering
        \includegraphics[width=0.09\textwidth]{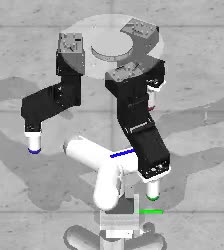}
        \includegraphics[width=0.09\textwidth]{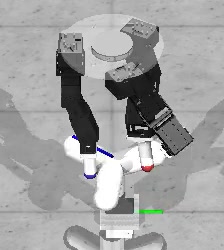}
        \includegraphics[width=0.09\textwidth]{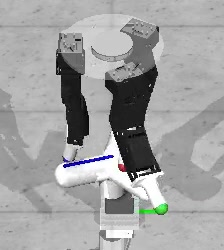}
        \includegraphics[width=0.09\textwidth]{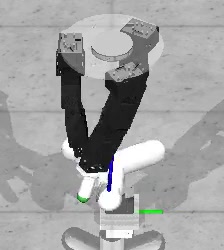}
        \includegraphics[width=0.09\textwidth]{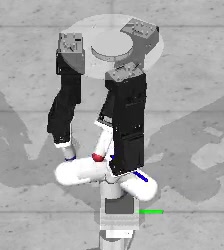}
    }
    \caption{We use off-DADS to enable skill discovery for D'Claw to manipulate the three-pronged valve. The first row represents a skill which turns the valve clockwise and the second row represents a skill which turns the valve counter-clockwise.}
    \label{fig:dclaw_skills}
\end{figure}
As we can see in Figure~\ref{fig:dclaw_skills}, the claw can learn to manipulate the valve and provide rotational skills in different directions. These skills can be used to manipulate the valve from random starting positions to random turn-goal positions using model-predictive control (MPC). The skills and model-predictive control for goal-turning of valve are available here:
\href{https://sites.google.com/view/dads-skill}{https://sites.google.com/view/dads-skill}
\end{document}